\pdfoutput=1

\PassOptionsToPackage{table}{xcolor}

\documentclass[11pt]{article}

\usepackage[final]{emnlp2023}

\usepackage{times}
\usepackage{latexsym}

\usepackage[T1]{fontenc}

\usepackage[utf8]{inputenc}

\usepackage{microtype}

\usepackage{inconsolata}

\usepackage{graphicx}
\usepackage{tabularx}

\usepackage[most]{tcolorbox}
\newtcolorbox{promptbox}[1][]{
  colback=gray!5,
  colframe=gray!50,
  fonttitle=\bfseries\small,
  title={#1},
  boxrule=0.5pt,
  arc=2pt,
  left=6pt,
  right=6pt,
  top=4pt,
  bottom=4pt,
  breakable
}
\definecolor{mygreen}{RGB}{34,120,50}
\definecolor{myred}{RGB}{178,34,52}
\newtcolorbox{querybox}[1][]{
  colback=gray!4, colframe=gray!45,
  fonttitle=\bfseries\small, coltitle=white, colbacktitle=gray!60,
  title={#1}, boxrule=0.4pt, arc=2pt,
  left=6pt, right=6pt, top=3pt, bottom=3pt, breakable
}
\newtcolorbox{gtbox}[1][]{
  colback=blue!3, colframe=blue!35,
  fonttitle=\bfseries\small, coltitle=white, colbacktitle=blue!55,
  title={#1}, boxrule=0.4pt, arc=2pt,
  left=6pt, right=6pt, top=3pt, bottom=3pt, breakable
}
\newtcolorbox{correctbox}[1][]{
  colback=mygreen!7, colframe=mygreen!55,
  fonttitle=\bfseries\small, coltitle=white, colbacktitle=mygreen!70,
  title={#1}, boxrule=0.4pt, arc=2pt,
  left=6pt, right=6pt, top=3pt, bottom=3pt, breakable
}
\newtcolorbox{wrongbox}[1][]{
  colback=myred!6, colframe=myred!50,
  fonttitle=\bfseries\small, coltitle=white, colbacktitle=myred!65,
  title={#1}, boxrule=0.4pt, arc=2pt,
  left=6pt, right=6pt, top=3pt, bottom=3pt, breakable
}
\newtcolorbox{neutralbox}[1][]{
  colback=purple!4, colframe=purple!35,
  fonttitle=\bfseries\small, coltitle=white, colbacktitle=purple!55,
  title={#1}, boxrule=0.4pt, arc=2pt,
  left=6pt, right=6pt, top=3pt, bottom=3pt, breakable
}
\newcommand{\good}[1]{\textcolor{mygreen}{\textbf{#1}}}
\newcommand{\bad}[1]{\textcolor{myred}{\textbf{#1}}}
\usepackage{pifont}
\newcommand{\cmark}{\ding{51}}
\newcommand{\xmark}{\ding{55}}
\usepackage{url}
\usepackage{caption}  
\usepackage{booktabs} 
\usepackage{wrapfig}
\usepackage{multirow}
\usepackage{amsmath}
\usepackage{amsfonts}
\usepackage{nicefrac}
\usepackage[table]{xcolor}
\usepackage{tikz}
\usetikzlibrary{positioning}
\usepackage{pgfplots}
\pgfplotsset{compat=1.18}

\title{AgentJudgeBench: A Multi-Difficulty Benchmark for Evaluating LLM Judges on Agentic Tool-Calling}

\author{Abhigya Verma \\ {\bf Amit Kumar Saha} \\ {\bf Seganrasan Subramanian} \\ {\bf Sai Harshitha Aluru} \\
        ServiceNow AI \\ Hyderabad, India}

\begin{document}

\maketitle

\begin{abstract}
LLM judges are widely used to evaluate agentic tool-calling systems, yet their reliability on structured, dependency-driven workflows remains largely unexamined. We present AgentJudgeBench, the first benchmark to systematically study LLM-as-a-judge reliability for agentic tool-calling over workflow DAGs, as distinct from the broader LLM-as-a-judge task of open-ended text or preference evaluation. The benchmark comprises 3,808 instances spanning six DAG topologies and three difficulty tiers, evaluated with five generators (3B--70B open-weight models and GPT-5.4) and six judges (20B to frontier scale) under paired with- and without-ground-truth conditions. Judge alignment degrades monotonically with task difficulty, 1.5$\times$ faster without ground truth, and on hard queries without ground truth all six judges converge to a narrow 77--82\% band regardless of scale, revealing a structural ceiling, driven primarily by task difficulty though its height is partly prompt-dependent for weaker generators, that model capacity alone cannot overcome. Ground-truth exposure is not uniformly beneficial: it reduces alignment for GPT-5.4 (1.5 pp) and Gemini-2.5-Pro (3.9 pp), consistent with over-anchoring. Among mitigation strategies, chain-of-thought reasoning and judge temperature both have negligible effect, while structured evaluation rubrics improve alignment by up to 6.5 pp but do not generalise uniformly across judge-generator pairs. With ground truth, QwQ-32B best matches the programmatic reference, while a human validation study identifies GPT-OSS-120B as the most human-aligned judge; without it, frontier judges lead only marginally within the shared ceiling. These results expose fundamental limitations of current LLM judges and yield practical guidelines for reliable evaluation in agentic systems. The code and dataset are available at \url{https://github.com/ServiceNow/SyGra/tree/scratch/agent_judge_bench/tasks/agentic_bfcl_judge_eval} and \url{https://huggingface.co/datasets/ServiceNow-AI/AgentJudgeBench}.
\end{abstract}
\vspace{-0.2cm}
\section{Introduction}
\vspace{-0.1cm}

\begin{figure*}[h!]
  \centering
\includegraphics[width=0.75\textwidth]{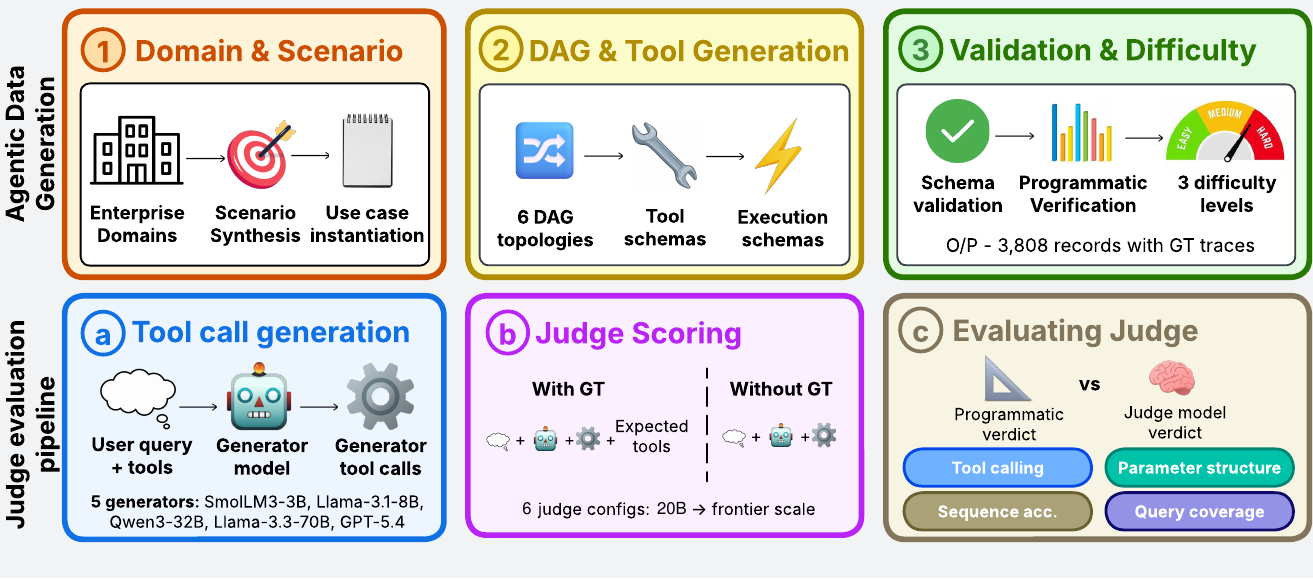} \vspace{-0.4cm} \caption{%
    Overall pipeline of AgentJudgeBench. Each BFCL-style Agentic record undergoes difficulty-controlled rewriting, generator inference, and parallel scoring by a programmatic judge and LLM judges (with/without GT), followed by alignment with the programmatic reference.%
  }
  \vspace{-0.3cm}
\label{fig:pipeline}
\end{figure*}

Using LLMs as automated judges has become standard practice for evaluating model outputs~\citep{zheng2023judging,tan2025judgebench,li2024arenahard}. On text-centric tasks -- dialogue, summarization, instruction following -- judge reliability is well characterized, with documented biases and known failure modes~\citep{zheng2023judging,wang2023fair}. As LLMs are increasingly deployed as autonomous agents that invoke tools and orchestrate multi-step workflows, the judge paradigm has been extended naturally to agentic tool-calling~\citep{qin2024toolllm,guo2024stabletoolbench,guo2025mcpagentbench}. This extension, however, has proceeded without a basic calibration check: \emph{how reliable are LLM judges in this structured setting?}

Agentic tool-calling differs from text evaluation in ways that matter for judge reliability. Correctness is not a matter of fluency or preference: it requires selecting the right tools from a typed schema, supplying well-formed arguments, ordering calls to respect execution dependencies, and covering all parts of the user's intent. A plan can fail in four orthogonal ways (tool selection, parameter structure, sequence accuracy, query coverage) that do not correlate cleanly -- a judge reliable at one may be blind to another, e.g., calibrated on simple sequential tasks yet fail on complex fan-in or diamond workflows where parallel branches must converge. When no ground-truth execution trace is available, the common deployment scenario, the judge must instead reconstruct correctness from the query and tool schemas alone, a fundamentally harder inference problem. Existing benchmarks that deploy LLM judges for tool-calling report only aggregate pass-rate agreement and vary none of these dimensions~\citep{qin2024toolllm,guo2024stabletoolbench,guo2025mcpagentbench}, leaving practitioners without a principled basis for choosing a judge, setting its configuration, or interpreting its outputs.

We introduce \textbf{AgentJudgeBench} to close this gap. The benchmark comprises 3,808 BFCL-style records~\citep{patil2025bfcl} spanning six DAG\footnote{A Directed Acyclic Graph (DAG) encodes execution dependencies between tool calls: nodes are tool invocations and directed edges indicate that one call must complete before another can begin, with the acyclicity constraint preventing circular dependencies.} topologies at three controlled difficulty tiers, each with a programmatically verified ground-truth trace. Five generators -- four open-weight (3B--70B) and one frontier (GPT-5.4) -- produce tool-calling outputs that are scored by six LLM judges (ranging from 20B open-weight to frontier closed systems) under paired with- and without-GT conditions across four structural metrics (tool selection, parameter structure, sequence accuracy, query coverage), yielding $321{,}648$ paired with-GT/without-GT evaluations (Appendix~\ref{app:eval-stats}). We organize our analysis around six research questions: which metrics and topologies are hardest (RQ1), how much judges agree with each other (RQ2), why without-GT alignment converges on hard queries (RQ3), and how sensitive alignment is to judge temperature (RQ4), chain-of-thought reasoning (RQ5), and prompt format (RQ6).

The results are non-obvious. Judge alignment degrades monotonically with query difficulty, $1.5\times$ faster without ground truth than with it, and all six judges converge to a $77$--$82\%$ ceiling on hard queries without a reference, regardless of model capacity; a follow-up ablation (\S\ref{sec:exp:ceiling}) confirms task difficulty as the primary driver for capable generators, though its height is somewhat prompt-dependent for weaker ones. GT exposure is not universally beneficial: two frontier judges (GPT-5.4, Gemini-2.5-Pro) are \emph{less} aligned when shown the reference, consistent with over-anchoring rather than independent judgement. Chain-of-thought adds at most $0.3$\,pp across 24 paired comparisons, and temperature has negligible effect (${\leq}0.25$\,pp spread, two judge-generator pairings). Structured per-metric prompt rubrics add $+4.8$--$+6.5$\,pp over free-form on one pairing, the largest lever we test, but a second pairing shows a smaller effect that reverses on hard queries, so we treat prompt format as impactful but judge/generator-dependent rather than universally dominant (\S\ref{sec:abl:format}). Inter-judge agreement is moderate ($\kappa \approx 0.42$ with GT), with systematic, capacity-correlated patterns not visible in aggregate scores.

\noindent\textbf{Scope.} We study LLM-as-a-judge specifically for agentic tool-calling, not the broader LLM-as-a-judge literature; DAG-structured tool-use data, synthetic dependency graphs, and programmatic trajectory scoring each have close prior work (\S2). Our contribution is the reliability \emph{protocol} that combines them -- paired with-GT/without-GT conditions, controlled difficulty, and per-metric decomposition -- applied to measure judge reliability rather than agent capability.

\noindent\textbf{Contributions.} \textbf{(1)}~A dataset of 3,808 records spanning six DAG topologies and three difficulty tiers with programmatically verified ground-truth traces, to be released publicly. \textbf{(2)}~A four-metric evaluation framework with a paired with-GT/without-GT protocol and bootstrap confidence intervals. \textbf{(3)}~A systematic six-RQ empirical study yielding actionable guidance: with ground truth, QwQ-32B best matches the programmatic reference, while GPT-OSS-120B is the most human-aligned judge in our validation study (Appendix~\ref{app:prog-judge-validation}); without ground truth, frontier judges lead narrowly, but the convergence ceiling limits the practical difference.

\section{Related Work}
Table~\ref{tab:overview} (Appendix~\ref{app:relwork-table}) surveys prior work across three related areas; we explain the key distinctions below.

\subsection{Agentic Data and Tool-Calling Benchmarks}
The dominant paradigm for evaluating tool-calling agents uses either environment-based execution feedback~\citep{zhou2024webarena,drouin2024workarena,trivedi2024appworld,yao2024taubench} or deterministic scoring against annotated trajectories~\citep{deng2023mind2web,xu2025agenttrek,patil2025bfcl}. $\tau$-bench~\citep{yao2024taubench} stresses agents against simulated users and proposes \texttt{pass\textasciicircum k} reliability rather than per-turn correctness; like this line generally, it scores agents but does not interrogate the judge producing those scores. BFCL~\citep{patil2025bfcl} is the closest structural ancestor to our data format, isolating single-turn function-calling with typed JSON schemas and AST-based ground-truth comparison; we adopt the same schema but extend it to multi-step DAG-structured workflows at three controlled difficulty tiers. The most structurally similar work is FuncBenchGen~\citep{maekawa2025funcbenchgen}, which frames multi-step calling as DAG traversal with controllable complexity. The key distinction is purpose: FuncBenchGen trains and evaluates \emph{generator} agents, whereas AgentJudgeBench measures \emph{judge} reliability on those generators' outputs via a paired with-GT/without-GT protocol and per-metric decomposition that FuncBenchGen does not provide. TaskBench~\citep{shen2024taskbench} shares the tool-dependency graph framing but focuses on decomposition quality rather than judge alignment. Synthetic data pipelines~\citep{wang2023selfinstruct,xu2025magpie,cui2024ultrafeedback} inform our generation approach but target model training rather than evaluation benchmarking.

\subsection{LLM-as-Judge}
\citet{zheng2023judging} established the LLM-as-judge paradigm on MT-Bench, showing GPT-4's strong agreement with human preferences on open-ended dialogue while identifying positional preference, verbosity sensitivity, and self-enhancement as systematic biases. Subsequent work either extends the evaluation surface -- JudgeBench~\citep{tan2025judgebench} and Arena-Hard-Auto~\citep{li2024arenahard} move to hard, verifiable response pairs -- or builds dedicated judge models~\citep{wang2024pandalm,li2024autoj,zhu2023judgelm,kim2024prometheus2}. JudgeLM~\citep{zhu2023judgelm} fine-tunes 7B--33B judges on GPT-4-distilled verdicts and identifies position, knowledge, and format biases that mirror those reported for prompt-based judges; Prometheus~2~\citep{kim2024prometheus2} adds rubric-conditioned direct assessment with open weights. Both target text-quality scoring rather than structural tool-calling correctness, but their bias taxonomies inform our prompt-format ablation (\S\ref{sec:abl:format}). The jury-of-judges paradigm~\citep{verga2024jury} shows that ensembling reduces individual judge biases. Crucially, this line targets text-centric tasks where judge quality reduces to preference alignment over fluent output. AgentJudgeBench occupies a different regime: correctness is structural, the verdict space is $\{0, 0.5, 1\}$ rather than a preference ranking, and failure modes are multi-dimensional and interdependent.

\vspace{-8pt}
\subsection{LLM Judges for Tool-Calling}
LLM judges have been deployed for tool-calling evaluation in several recent benchmarks, but always as an implementation detail rather than the subject of study. ToolLLM~\citep{qin2024toolllm} uses ChatGPT to compute pass rates over API trajectories, reporting 87.1\% human agreement in aggregate; StableToolBench~\citep{guo2024stabletoolbench} and MCP-AgentBench~\citep{guo2025mcpagentbench} adopt similar approaches with updated judge models. GeoBenchX~\citep{krechetova2025geobenchx} assembles a three-judge panel achieving 88--96\% agreement, and Agent-as-a-Judge~\citep{zhuge2024agentjudge} and Auto-Eval Judge~\citep{bhonsle2025autoeval} propose more structured evaluation frameworks. In each case, judge reliability is reported as a single aggregate figure on a small sample, with no variation of task complexity, difficulty, or ground-truth availability. ToolSandbox~\citep{lu2024toolsandbox} notably questions LLM judge reliability directly but replaces judges with programmatic evaluation rather than characterising their failure modes. AgentJudgeBench takes the complementary stance: keep the judge paradigm and measure it systematically, decomposing reliability by metric, topology, difficulty, and condition across $321{,}648$ completed evaluations (Appendix~\ref{app:eval-stats}). A detailed feature comparison across all seven related systems appears in Appendix~\ref{app:judge-comparison}.

\section{Methodology}
\label{sec:methodology}

Figure~\ref{fig:pipeline} gives an end-to-end view of the pipeline. Each record is expanded into three difficulty variants; every generator $g \in \mathcal{G}$ produces tool-call predictions; a deterministic programmatic judge scores each prediction to yield a reference vector $\mathbf{p}_r$; and every LLM judge $j \in \mathcal{J}$ produces paired verdicts $\boldsymbol{\ell}^{\text{GT}}_{j,r}$, $\boldsymbol{\ell}^{\text{without GT}}_{j,r}$, compared against $\mathbf{p}_r$ via Eq.~\ref{eq:alignment}. The full pipeline is implemented as computational graphs on \textbf{SyGra}~\citep{pradhan2025sygra}, an open-source graph-oriented synthetic-data-generation framework.

\vspace{-0.2cm}
\subsection{Data Generation}
\label{sec:methodology:data}
\vspace{-0.1cm}
The benchmark's BFCL-style agentic data is constructed through a multi-stage synthetic pipeline that generates complete agentic records from minimal seed inputs.

\noindent\textbf{Why synthetic.} We generate records synthetically rather than mining real enterprise traces because the two are not interchangeable for this study's purpose: a controlled reliability study needs, for every record, a ground-truth trace we can certify as correct; systematic coverage across six DAG topologies and three difficulty tiers rather than whatever distribution occurs in logs; and enough scale and domain diversity (15 enterprise domains) to isolate structural effects from domain idiosyncrasy. Real enterprise traces satisfying all three properties are difficult to source, requiring proprietary internal tool schemas, live or replayable execution environments, and an independent way to certify trace correctness -- rarely available outside a single organisation's internal systems. Generating synthetically lets us guarantee a verified reference trace for every one of the $3{,}808$ records and control topology and difficulty independently, which is what makes the paired with-GT/without-GT protocol (\S\ref{sec:methodology:eval}) possible; the Limitations section discusses what this trades off (domain drift, interactive multi-turn execution) and the validation evidence for this design choice.

\noindent\textbf{Record construction.} Given an enterprise domain label (e.g., IT service management, contract lifecycle management, energy grid operations), the pipeline generates use-case scenarios, synthesizes a typed tool inventory as function signatures with constrained return types (\texttt{str}, \texttt{bool}, \texttt{list}, \texttt{dict}, \texttt{None}), and produces executable pseudocode linking all tool dependencies. A natural language user utterance is then paired with function input/output specifications and formalized into a JSON schema with typed arguments, required fields, and output definitions; finally, an ordered execution trace captures sequential and parallel tool calls, their inputs, outputs, and step descriptions.

\noindent\textbf{DAG topologies.} Records are organized into six topologies derived from common dependency patterns in real enterprise agentic workflows -- \textit{linear}, \textit{fan-out}, \textit{fan-in}, \textit{diamond}, \textit{optional enrichment}, and \textit{loop-like} -- from single-path execution to iterative loops.

\noindent\textbf{Difficulty tiers.} Each record is expanded into three levels (\textit{easy}, \textit{medium}, \textit{hard}) by increasing query ambiguity while holding the task structure and GT trace constant. Rewrite validation (Appendix~\ref{app:rewrite-validation}) confirms the medium$\to$hard step on $93.9\%$ of records; the easy$\to$medium step is unanimously validated on only $58.1\%$, and in roughly $41.9\%$ of cases medium is better characterised as a paraphrase than a strict difficulty increase. The primary difficulty-degradation evidence therefore comes from the medium$\to$hard step; medium data should be read as a robustness check rather than a calibrated mid-point (see Limitations).

\noindent\textbf{Validation.} GT traces pass a two-level quality gate: each record is first structurally verified via JSON-schema validation, argument type checking, and trace consistency checks, then each (utterance, tool-call) pair is programmatically checked for argument sufficiency, grounding alignment, and query naturalness, with failing records regenerated. No model from $\mathcal{J}$, or any other LLM, is involved in either gate check, ruling out self-reinforcing bias. A 120-record human annotation study on hard-difficulty records (Appendix~\ref{app:prog-judge-validation}), stratified across all six topologies, validates the scorer at $92.7\%$ metric-level agreement with independent human judgement (scope and limitations in the Limitations section). The final corpus comprises $3{,}808$ records spanning 15 enterprise domains with 8--19 tools per record, across six topologies and three difficulty levels.

\vspace{-0.4cm}
\subsection{Evaluation}
\label{sec:methodology:eval}
\begin{figure}[t]
  \centering
  \includegraphics[width=0.95\columnwidth]{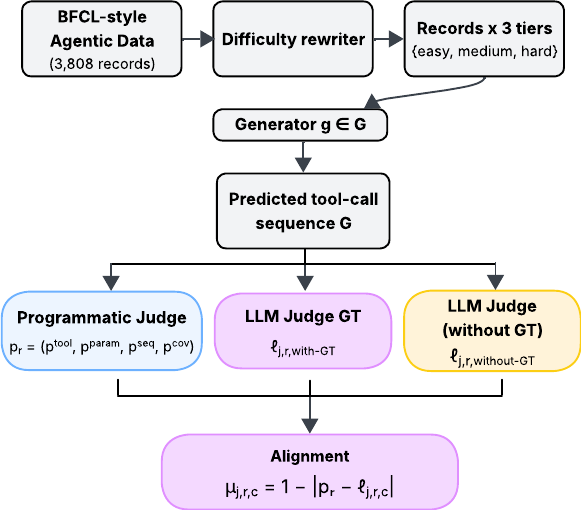} \vspace{-0.1cm}
  \caption{Evaluation pipeline of AgentJudgeBench. Records expand into three difficulty tiers, LLM judges (with/without GT) and programmatic judge score the generated tool-call predictions.}
  \label{fig:eval}
\vspace{-0.4cm}
\end{figure}

\vspace{-0.3cm}
As shown in Figure~\ref{fig:eval}, our evaluation is layered: a deterministic \emph{programmatic judge} computes a reference score on every record, an \emph{LLM judge} produces a per-metric verdict on the same record, and an \emph{alignment aggregator} compares the two. Each layer is defined below; symbol definitions appear in Appendix~\ref{app:notation}, and all prompts are reproduced verbatim in Appendix~\ref{app:prompts}.

\noindent\textbf{Programmatic Judge}
\label{sec:methodology:programmatic}

\noindent\textbf{Why a deterministic reference.} The programmatic judge, not a human or LLM annotator, serves as the reference signal against which every LLM judge (and, transitively, the generator) is measured: using another LLM would reintroduce the same reliability question one level up, and a human-in-the-loop reference cannot scale to the $321{,}648$ evaluations in the full factorial grid, whereas a deterministic, rule-based scorer gives every cell an identical, reproducible reference. We validate rather than assume this choice: ground-truth traces pass a two-level programmatic quality gate with no LLM involved (below), and a 120-record independent human study (Appendix~\ref{app:prog-judge-validation}) directly checks where the scorer's notion of correctness agrees and disagrees with human judgement. Let $G = (g_1, \dots, g_{|G|})$ denote the generator's predicted ordered sequence of tool calls and $E = (e_1, \dots, e_{|E|})$ the GT ordered sequence. Let $\mathrm{name}(\cdot)$ return a tool's identifier, $\mathrm{args}(\cdot)$ its argument-key set, and let $\mathcal{G} = \{\mathrm{name}(g_i)\}$, $\mathcal{E} = \{\mathrm{name}(e_i)\}$ denote the unordered sets of tool identifiers. The programmatic judge emits four independent per-record scores in $[0, 1]$. Throughout the paper we call each of these four scores -- tool selection, parameter structure, sequence accuracy, query coverage -- a \emph{metric}: the term denotes one axis of tool-calling correctness scored independently by both the programmatic judge and the LLM judge, not a distinct evaluation instrument or benchmark-level metric.

\textbf{Tool-selection accuracy $p^{\text{tool}}$ and sequence accuracy $p^{\text{seq}}$.}
$p^{\text{tool}}$ penalises set-level mismatch; $p^{\text{seq}}$ measures position-by-position identity, normalised by expected length:

\vspace{-0.4cm}
\begin{subequations}
\small
\begin{align}
p^{\text{tool}}
&=
\max\!\left(
0,\,
1
-
\frac{
|\mathcal{G} \setminus \mathcal{E}|
+
|\mathcal{E} \setminus \mathcal{G}|
}{
\max(|\mathcal{E}|,1)
}
\right)
\label{eq:prog-tool}
\\[6pt]
p^{\text{seq}}
&=
\frac{1}{\max(|E|,1)}
\sum_{i=1}^{\min(|G|,|E|)}
\mathbb{1}\!\left[
\mathrm{name}(g_i)
=
\mathrm{name}(e_i)
\right]
\label{eq:prog-seq}
\end{align}
\vspace{-0.2cm}
\end{subequations}

\noindent with $p^{\text{tool}} = p^{\text{seq}} = 1$ when the respective sequences are empty.

\noindent\textbf{Parameter-structure accuracy $p^{\text{param}}$.}
For each predicted call $g \in G$ let $A_g = \mathrm{args}(g)$ and let $A^{*}_{g}$ be the expected-argument key set of the call in $E$ with the same tool identifier (undefined if the predicted tool is not in $\mathcal{E}$). Define per-call structural scores
\vspace{-0.3cm}
\begin{equation}
\small
\begin{aligned}
s(g)=
\begin{cases}
1.0
& \text{if } A_g = A^{*}_{g}
\\[-1pt]
& \quad \text{(all keys present, no extras)}
\\[4pt]

0.5
& \text{if } A^{*}_{g} \subseteq A_g
\\[-1pt]
& \quad \text{and } A_g \setminus A^{*}_{g} \neq \emptyset
\\[-1pt]
& \quad \text{(extras only)}
\\[4pt]

0.0
& \text{if } A^{*}_{g} \not\subseteq A_g
\\[-1pt]
& \quad \text{or } A^{*}_{g}\ \text{undefined}
\end{cases}
\end{aligned}
\label{eq:prog-param-per-call}
\end{equation}
and aggregate by mean: $p^{\text{param}} = \bigl(\sum_{g \in G} s(g)\bigr) / \max(|G|, 1)$, with $p^{\text{param}} = 1$ when $G$ is empty.

\noindent\textbf{Query-coverage accuracy $p^{\text{cov}}$.}
The fraction of expected tool identifiers covered by the prediction, ignoring extras:
\begin{equation}
p^{\text{cov}} \;=\;
\begin{cases}
  1                                        & \text{if } \mathcal{E} = \emptyset, \\
  0                                        & \text{if } \mathcal{G} \cap \mathcal{E} = \emptyset, \\
  \lvert \mathcal{G} \cap \mathcal{E} \rvert / \lvert \mathcal{E} \rvert & \text{otherwise}.
\end{cases}
\label{eq:prog-cov}
\end{equation}
\noindent\textbf{Record-level aggregate.}
Each record receives a four-element programmatic vector $\mathbf{p}_r = (p^{\text{tool}}_r, p^{\text{param}}_r, p^{\text{seq}}_r, p^{\text{cov}}_r)$. Where a scalar is needed, specifically in the generator-accuracy reporting of Table~\ref{tab:full-results}, we use the unweighted mean $\bar{p}_r = \frac{1}{M} \sum_{m=1}^{M} p^m_r$, where $M=4$ is the number of metrics; elsewhere the four components are kept separate and averaged independently. We adopt equal weights since the four metrics target orthogonal aspects of tool-call correctness with no principled basis for preferring one; Section~\ref{sec:exp:main} gives per-metric breakdowns practitioners can reweight for a given application (e.g., upweighting sequence accuracy for strict-ordering workflows).
Empirically, the judge ranking is stable under alternative metric weightings: Spearman $\rho \geq 0.83$\footnote{Spearman's $\rho$ is a rank correlation coefficient measuring the monotone association between two orderings; $\rho=1$ indicates identical rankings, $\rho=0$ no association, and $\rho=-1$ perfectly reversed rankings.} between the equal-weight ranking and five alternative schemes (seq${\times}2$, param${\times}2$, cov${\times}0.5$, seq${\times}2{+}$param${\times}2$), confirming that the equal-weight aggregate is an adequate proxy for any practitioner-specific weighting.

\noindent\textbf{LLM Judge}
\label{sec:methodology:llmjudge}
The same generator outputs scored by the programmatic judge are independently evaluated by every LLM judge $j \in \mathcal{J}$. Each judge receives a structured prompt containing the original user query, the full set of available tool schemas, and the generator's predicted tool-call sequence, and is asked to produce a single JSON object scoring each of the four metrics (tool selection, parameter structure, sequence accuracy, query coverage) on a $\{0, 0.5, 1\}$ scale, along with a one-sentence justification per metric and an overall assessment.

Two prompt variants exist, corresponding to the conditions $c \in \{\text{with-GT}, \text{without-GT}\}$: the \textbf{with-GT} prompt additionally exposes the GT tool-call sequence as a reference block, while the \textbf{without-GT} prompt omits it entirely, requiring the judge to assess correctness from the query and tool schemas alone. Judge decoding is held constant across all $(g, j, d, c)$ configurations to isolate the effect of our independent variables. The full prompt bodies appear in Appendix~\ref{app:prompts}.

\noindent\textbf{Alignment Against the Programmatic Reference}
\label{sec:methodology:alignment}
Given a record $r$, condition $c$, judge $j$, and metric $m \in \{\text{tool}, \text{param}, \text{seq}, \text{cov}\}$, let $\ell^{m}_{j,r,c} \in \{0, 0.5, 1\}$ denote the judge's verdict and $p^{m}_r$ the programmatic value. The per-metric match score and record-level aggregate are:
\vspace{-0.2cm}
\begin{subequations}
\begin{align}
\mu^{m}_{j,r,c}
&=
1 - \left| p^{m}_r - \ell^{m}_{j,r,c} \right|
\label{eq:per-metric-match}
\\[4pt]
\mu_{j,r,c}
&=
\frac{1}{M}
\sum_{m=1}^{M}
\mu^{m}_{j,r,c}
\label{eq:overall_alignment}
\end{align}
\end{subequations}
\noindent exact match on all metrics yields $\mu = 1$ and maximally divergent verdict yields $0$. The configuration-level alignment percentage and GT lift are:
\vspace{-0.2cm}
\begin{subequations}
\begin{align}
\mathrm{align}(j, g, d, c)
&=
\frac{100}{N_{g,d}}
\sum_{r=1}^{N_{g,d}}
\mu_{j,r,c}
\label{eq:alignment}
\\[4pt]
\mathrm{lift}(j)
&=
\overline{\mathrm{align}}_{\text{GT}}(j)
-
\overline{\mathrm{align}}_{\text{without GT}}(j)
\label{eq:lift}
\end{align}
\end{subequations}
\noindent where the overline denotes the mean over all 12 $(g, d)$ configurations. Perfect reproduction of the programmatic vector yields $\mathrm{align} = 100\%$; an independent judge attains $50\%$ in expectation.

In the experiments, we report both aggregate alignment $\mathrm{align}(j, g, d, c)$ and per-metric breakdowns to identify which dimensions of tool-calling correctness judges find most difficult to assess.
\vspace{-0.2cm}
\section{Experiments}
\label{sec:experiments}
\vspace{-0.15cm}
We use AgentJudgeBench to evaluate the extent to which LLM-judge alignment depends on the generator, query difficulty, and ground-truth availability. Protocol details (prompts, scoring, conditions) are in Section~\ref{sec:methodology}; model identifiers are in Appendix~\ref{app:generators} and Appendix~\ref{app:judges}.

\subsection{Experimental Setup}
\label{sec:exp:design}
We instantiate a fully-crossed factorial over three factors: generator $g \in \mathcal{G}$ (five models, 3B to frontier), judge $j \in \mathcal{J}$ (six LLMs), and difficulty $d \in \{\text{easy}, \text{medium}, \text{hard}\}$. Each cell is observed under both with GT and without GT conditions, yielding 90 factorial cells and $321{,}648$ valid (generator, judge, difficulty, record) tuples. Exact per-cell counts are in Appendix~\ref{app:eval-stats}. We report $\mathrm{align}(j, g, d, c)$ (Eq.~\ref{eq:alignment}) and, where relevant, GT lift (Eq.~\ref{eq:lift}).

\subsection{Results and Analysis}
\label{sec:exp:main}
Table~\ref{tab:full-results} reports every configuration in the factorial design. Each generator occupies a sub-block with six rows corresponding to the judges in $\mathcal{J}$, plus a seventh row for Prometheus-2~\citep{kim2024prometheus2}, a judge-specialised model added as a baseline across all five generators and both GT conditions (Limitations), shown for reference and excluded from the six-judge statistics below. Columns pair each difficulty level with its with-GT and without-GT conditions; bold entries mark the highest alignment within each (difficulty, condition) column of each sub-block.

\textbf{Finding 1: Monotone difficulty degradation.} All 30 (generator, judge) pairs exhibit strictly monotone alignment degradation from easy to hard under both conditions (Table~\ref{tab:full-results}), without-GT degradation roughly $1.5\times$ larger than with-GT. On hard without-GT records, all six judges converge to a narrow $77$--$82\%$ band across four of five generators (including frontier GPT-5.4), indicating a task-level rather than judge-level ceiling; degradation curves are in Appendix~\ref{app:degradation}.
\begin{table}[h]
\centering
\small
\setlength{\tabcolsep}{3pt}
\renewcommand{\arraystretch}{1.02}

\resizebox{\columnwidth}{!}{%
\begin{tabular}{llrrrrr}
\toprule

\textbf{Judge} &
\textbf{Diff} &
\textbf{GT} &
\textbf{withoutGT} &
\textbf{cGT} &
\textbf{$\Delta_{\text{cGT-GT}}$} &
\textbf{$\Delta_{\text{cGT-noGT}}$} \\

\midrule

\multirow{3}{*}{QwQ-32B}
& Easy   & 96.6 & 95.6 & 95.5 & $-1.1$ & $-0.1$ \\
& Medium & 92.3 & 92.0 & 92.0 & $-0.3$ & $0.0$ \\
& Hard   & 84.6 & 84.4 & 84.6 & $0.0$ & $+0.2$ \\

\addlinespace[2pt]

\multirow{3}{*}{Gemini}
& Easy   & 90.7 & 95.7 & 92.5 & $+1.8$ & $-3.2$ \\
& Medium & 86.4 & 92.1 & 86.4 & $0.0$ & $-5.7$ \\
& Hard   & 78.9 & 84.6 & 78.9 & $0.0$ & $-5.7$ \\

\bottomrule
\end{tabular}%
}
\vspace{-0.1cm}
\caption{Full C3 results on Llama-3.3-70B generator. All values are mean alignment (\%) over 3,764--3,771 records per cell.
\textbf{cGT}: corrupted-GT condition. \textbf{Gemini}: Gemini 2.5 Pro.
\textbf{$\Delta_{\text{cGT-GT}}$}: corrupted-GT minus standard GT.
\textbf{$\Delta_{\text{cGT-noGT}}$}: corrupted-GT minus without GT.}
\vspace{-0.2cm}
\label{tab:c3-anchoring}

\end{table}
\textbf{Scope of Finding 1.} The medium$\to$hard drop (${\approx}5$--$7$\,pp with GT) is the primary evidence for difficulty-driven degradation; hard rewrites are unanimously validated on $93.9\%$ of records (Appendix~\ref{app:rewrite-validation}), while the easy$\to$medium step (unanimous on $58.1\%$) partially reflects paraphrase-robustness.

\begin{figure*}[t]
\centering
  \includegraphics[width=0.85\textwidth]{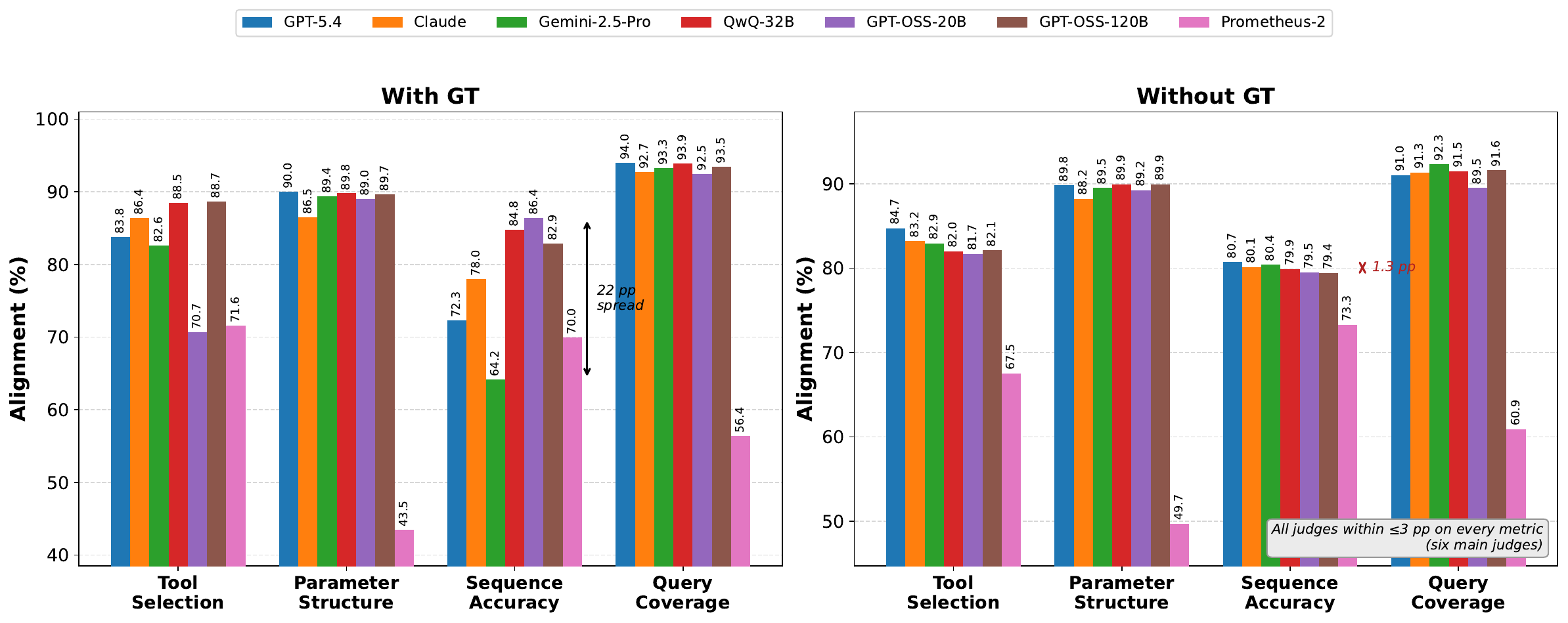}
  \vspace{-0.2cm}
  \caption{Per-metric alignment (with and without GT) across the six main judges, plus Prometheus-2 (judge-specialised baseline; Limitations), shown for reference.}
  \label{fig:permetric}
  \vspace{-0.5cm}
\end{figure*}

\textbf{Finding 2: Ground-truth exposure is not monotonically beneficial.} GT lift (Eq.~\ref{eq:lift}) is positive for QwQ-32B and GPT-OSS-120B, but negative for GPT-5.4 and Gemini-2.5-Pro (non-overlapping bootstrap CIs; Appendix~\ref{app:bootstrap-ci}). Figure~\ref{fig:permetric} shows the effect concentrates in sequence accuracy: frontier judges anchor to the GT trace's ordering and penalise functionally equivalent but structurally deviant sequences (case studies in Appendix~\ref{app:anchoring-cases}).
\begin{table}[t]
\centering

\small
\setlength{\tabcolsep}{2.4pt}
\renewcommand{\arraystretch}{0.80}

\begin{tabular*}{\columnwidth}{@{\extracolsep{\fill}}llcccccc}

\toprule

& &
\multicolumn{2}{c}{\scriptsize\textbf{Easy}} &
\multicolumn{2}{c}{\scriptsize\textbf{Med.}} &
\multicolumn{2}{c}{\scriptsize\textbf{Hard}} \\

\cmidrule(lr){3-4}
\cmidrule(lr){5-6}
\cmidrule(lr){7-8}

\scriptsize\textbf{Gen.} &
\scriptsize\textbf{Judge} &
\scriptsize\textbf{GT} &
\scriptsize\textbf{No GT} &
\scriptsize\textbf{GT} &
\scriptsize\textbf{No GT} &
\scriptsize\textbf{GT} &
\scriptsize\textbf{No GT} \\

\midrule

\multirow{7}{*}{\rotatebox{90}{\scriptsize\textit{Sm3B}}}
& OSS20  & 84.8 & 85.3 & 81.7 & 79.3 & 77.4 & 69.0 \\
& QwQ32  & \textbf{90.0} & 86.7 & \textbf{84.8} & 80.3 & 77.2 & 69.7 \\
& OSS120 & 89.2 & 86.3 & \textbf{84.8} & 80.2 & \textbf{77.8} & 69.7 \\
& ClS4   & 86.5 & 86.8 & 81.3 & 80.9 & 74.2 & 71.2 \\
& Gem    & 81.1 & 86.6 & 77.7 & 80.7 & 73.4 & 70.8 \\
& GPT5   & 86.0 & \textbf{87.3} & 81.6 & \textbf{82.0} & 76.9 & \textbf{73.3} \\
& Prom2  & 64.0 & 67.1 & 61.6 & 64.9 & 58.6 & 60.2 \\

\midrule

\multirow{7}{*}{\rotatebox{90}{\scriptsize\textit{Ll8B}}}
& OSS20  & 87.4 & 89.5 & 85.3 & 85.7 & 81.2 & 75.9 \\
& QwQ32  & \textbf{94.2} & 91.2 & \textbf{90.8} & 86.8 & \textbf{84.1} & 76.6 \\
& OSS120 & 93.1 & 90.7 & 89.8 & 86.6 & 83.6 & 76.6 \\
& ClS4   & 90.3 & 90.6 & 86.5 & 86.6 & 80.3 & 77.1 \\
& Gem    & 85.9 & \textbf{91.9} & 83.3 & \textbf{87.6} & 78.7 & 78.2 \\
& GPT5   & 88.5 & 91.4 & 85.0 & 87.5 & 80.6 & \textbf{79.0} \\
& Prom2  & 60.8 & 63.2 & 59.0 & 60.8 & 56.7 & 58.4 \\

\midrule

\multirow{7}{*}{\rotatebox{90}{\scriptsize\textit{Qw32B}}}
& OSS20  & 88.4 & 93.2 & 85.8 & 88.7 & 82.7 & 81.1 \\
& QwQ32  & \textbf{95.0} & 93.9 & \textbf{90.9} & 89.1 & \textbf{84.7} & 81.1 \\
& OSS120 & 93.9 & 94.0 & 90.0 & 89.2 & 84.4 & 81.3 \\
& ClS4   & 92.7 & 94.0 & 88.5 & 89.2 & 82.1 & 81.6 \\
& Gem    & 86.4 & \textbf{94.1} & 83.0 & 89.2 & 78.1 & 81.4 \\
& GPT5   & 90.5 & 93.9 & 86.4 & \textbf{89.4} & 81.9 & \textbf{81.7} \\
& Prom2  & 64.4 & 66.7 & 62.0 & 64.5 & 59.1 & 61.5 \\

\midrule

\multirow{7}{*}{\rotatebox{90}{\scriptsize\textit{Ll70B}}}
& OSS20  & 89.0 & 94.5 & 87.2 & 91.2 & 83.9 & 84.0 \\
& QwQ32  & \textbf{96.6} & 95.6 & \textbf{93.6} & 92.1 & \textbf{87.5} & 84.4 \\
& OSS120 & 96.0 & 95.5 & 93.0 & 92.1 & 87.2 & 84.5 \\
& ClS4   & 93.3 & 94.6 & 89.9 & 90.7 & 83.7 & 83.3 \\
& Gem    & 90.5 & \textbf{95.7} & 86.5 & \textbf{92.2} & 82.2 & \textbf{84.6} \\
& GPT5   & 91.4 & 95.2 & 87.4 & 91.7 & 83.0 & \textbf{84.6} \\
& Prom2  & 66.5 & 68.6 & 62.8 & 65.2 & 58.3 & 60.2 \\

\midrule

\multirow{7}{*}{\rotatebox{90}{\scriptsize\textit{GPT5}}}
& OSS20  & 83.1 & 85.6 & 82.1 & 83.9 & 79.4 & 79.5 \\
& QwQ32  & 85.7 & 85.8 & 84.3 & 84.2 & 80.5 & 79.8 \\
& OSS120 & 84.8 & 85.3 & 83.9 & 83.3 & 80.7 & 78.6 \\
& ClS4   & \textbf{86.1} & \textbf{87.3} & \textbf{84.5} & \textbf{85.4} & 80.7 & \textbf{81.2} \\
& Gem    & 81.6 & 85.2 & 80.2 & 83.3 & 77.6 & 78.4 \\
& GPT5   & 85.8 & 87.0 & 84.1 & 84.9 & \textbf{81.4} & 79.8 \\
& Prom2  & 57.1 & 60.9 & 57.9 & 61.1 & 57.8 & 60.5 \\

\bottomrule
\end{tabular*}
\vspace{-0.2cm}
\caption{LLM--judge alignment (\%) with the programmatic reference across generators, judges, difficulty, and GT condition. Bold = column-wise maximum within each generator block among the six main judges. Prom2 = Prometheus-2, a judge-specialised baseline (Limitations); shown for reference and not counted toward bold column-wise maxima or six-judge statistics.}
\label{tab:full-results}
\vspace{-0.7cm}
\end{table}
\textbf{C3 control.} Replacing the reference with a wrong GT from a different record confirms pure anchoring (Table~\ref{tab:c3-anchoring}): Gemini-2.5-Pro aligns identically under standard and corrupted GT, while QwQ-32B tracks without-GT within $0.2$\,pp. Full results are in Appendix~\ref{app:c3-table}.
\footnotetext{Cohen's $\kappa$ measures inter-rater agreement corrected for chance: ($0$=chance, $1$=perfect; $0.2$--$0.4$ ``fair,'' $0.4$--$0.6$ ``moderate,'' ${\geq}0.6$ ``substantial'') agreement. All $\kappa$ values here fall in the fair-to-substantial range ($0.225$--$0.606$), consistent with the structural complexity of the four-metric scoring task.}
\textbf{Finding 3: Best judge depends on configuration.} QwQ-32B leads GT alignment in 10 of 15 (generator, difficulty) cells but is never the top without-GT judge; Gemini-2.5-Pro and GPT-5.4 lead without-GT on stronger and weaker generators, respectively. See Appendix~\ref{app:decision-guide} for a deployment decision table. The following research questions explain the mechanisms behind Findings 1--3.

\textbf{RQ1: Which metrics and DAG topologies are hardest for judges?}
\label{sec:exp:disaggregation}

Figure~\ref{fig:permetric} breaks alignment into per-metric components: with GT, sequence accuracy is the weakest dimension with substantial inter-judge spread, while without GT all judges compress to a narrow band, reflecting the $1.0$-default rubric. DAG topology imposes a judge-independent difficulty ordering (\emph{fan-out} easiest, \emph{loop-like}/\emph{fan-in} hardest); full breakdowns are in Appendices~\ref{app:permetric-table}--\ref{app:pertopology-table}.

\textbf{RQ2: How much do judges agree with each other?}
\label{sec:exp:interjudge}

Table~\ref{tab:interjudge} reports pairwise judge agreement. Mean agreement is $79.1\%$ ($\kappa=0.419$) with GT and $92.6\%$ ($\kappa=0.559$) without GT; the higher without-GT value reflects prompt-driven verdict compression rather than genuine consensus. The highest agreement is QwQ-32B$\times$GPT-OSS-120B ($89.4\%$, $\kappa=0.606$), and the lowest is Claude Sonnet 4.5$\times$GPT-OSS-20B ($70.4\%$, $\kappa=0.225$). Under without-GT, pairwise $\kappa$ decreases monotonically with judge tier separation ($\rho=-0.825$\footnote{Spearman's $\rho$ measures the rank correlation between the pairwise judge tier gap (by parameter count) and their $\kappa$ agreement score; $|\rho|>0.7$ indicates a strong monotone relation.}, $p<0.01$); this relationship vanishes under GT ($\rho=-0.171$), where GT acts as a shared anchor. Disagreement concentrates at the $0.5$ partial-credit boundary ($94$--$97\%$ of off-diagonal entries); verdict-level confusion matrices are in Appendix~\ref{app:confusion} and a binary scale ablation is in Appendix~\ref{app:abl-binary}. Prometheus-2 agrees with every judge at close to chance level under both conditions ($42$--$52\%$ exact, $\kappa = 0.01$--$0.07$; Table~\ref{tab:interjudge}), well below the fair-to-substantial range spanned by the six main judges, indicating it forms its own bloc rather than a lower-alignment member of the general-purpose cluster (Limitations).

\begin{table}[t]
\centering
\small
\setlength{\tabcolsep}{3.5pt}
\renewcommand{\arraystretch}{1.02}

\resizebox{\columnwidth}{!}{%
\begin{tabular}{lccccccc}
\toprule

& \textbf{OSS-20B}
& \textbf{QwQ-32B}
& \textbf{OSS-120B}
& \textbf{Claude S. 4.5}
& \textbf{Gemini}
& \textbf{GPT-5.4}
& \textbf{Prom2} \\

\midrule

OSS-20B
& {-}
& \cellcolor{blue!8}77.4
& \cellcolor{blue!8}78.4
& \cellcolor{blue!8}70.4
& \cellcolor{blue!8}74.5
& \cellcolor{blue!8}75.2
& \cellcolor{blue!8}42.2 \\

QwQ-32B
& \cellcolor{orange!15}0.386
& {-}
& \cellcolor{blue!8}\textbf{89.4}
& \cellcolor{blue!8}85.1
& \cellcolor{blue!8}80.7
& \cellcolor{blue!8}79.7
& \cellcolor{blue!8}49.0 \\

OSS-120B
& \cellcolor{orange!15}0.414
& \cellcolor{orange!15}\textbf{0.606}
& {-}
& \cellcolor{blue!8}83.2
& \cellcolor{blue!8}81.6
& \cellcolor{blue!8}79.8
& \cellcolor{blue!8}48.0 \\

Claude S. 4.5
& \cellcolor{orange!15}0.225
& \cellcolor{orange!15}0.490
& \cellcolor{orange!15}0.424
& {-}
& \cellcolor{blue!8}76.5
& \cellcolor{blue!8}78.4
& \cellcolor{blue!8}47.4 \\

Gemini
& \cellcolor{orange!15}0.387
& \cellcolor{orange!15}0.411
& \cellcolor{orange!15}0.437
& \cellcolor{orange!15}0.328
& {-}
& \cellcolor{blue!8}76.4
& \cellcolor{blue!8}44.1 \\

GPT-5.4
& \cellcolor{orange!15}0.419
& \cellcolor{orange!15}0.446
& \cellcolor{orange!15}0.449
& \cellcolor{orange!15}0.430
& \cellcolor{orange!15}0.432
& {-}
& \cellcolor{blue!8}42.6 \\

Prom2
& \cellcolor{orange!15}0.026
& \cellcolor{orange!15}0.074
& \cellcolor{orange!15}0.055
& \cellcolor{orange!15}0.062
& \cellcolor{orange!15}0.009
& \cellcolor{orange!15}0.036
& {-} \\

\bottomrule
\end{tabular}%
}
\vspace{-0.2cm}
\caption{Pairwise inter-judge agreement under the with GT condition.
\colorbox{blue!12}{\strut Upper triangle}: exact metric-level agreement (\%);
\colorbox{orange!20}{\strut Lower triangle}: Cohen's $\kappa$\protect\footnotemark. Prom2 (Prometheus-2, judge-specialised baseline) row/column added for reference; excluded from the six-judge mean agreement statistics discussed in the text.}
\label{tab:interjudge}
\vspace{-0.3cm}
\end{table}

A soft jury of all six judges matches but does not exceed the top individual judge ($82.5\%$ with GT hard)~\citep{verga2024jury}; ensemble and individual rankings against human annotators are in Appendix~\ref{app:prog-judge-validation}.

\textbf{RQ3: Why does without-GT alignment converge on hard queries?}
\label{sec:exp:ceiling}

The $77$--$82\%$ hard without-GT convergence (Table~\ref{tab:full-results}) has three consistent explanations: (1) the without-GT prompt defaults to $1.0$, suppressing hard-query discrimination; (2) generator error peaks on hard records, compressing all judges identically; and (3) per-metric compression is uniform (Table~\ref{tab:e4-permetric}). The six-judge ensemble achieves $79.5\%$, matching the best individual judge (GPT-5.4, $79.8\%$) to within $0.4$\,pp, confirming correlated, structural failure rather than independent per-judge noise.

\begin{table}[t]
\centering
\small
\setlength{\tabcolsep}{5pt}
\renewcommand{\arraystretch}{1.02}

\resizebox{\columnwidth}{!}{%
\begin{tabular}{lccc}
\toprule

\textbf{Generator} &
\textbf{1.0-default} &
\textbf{0.5-default} &
\textbf{$\Delta$} \\

\midrule

Llama-3.3-70B & 84.3 & 84.7 & $+0.4$ \\
Qwen3-32B     & 81.4 & 82.4 & $+1.0$ \\
GPT-5.4       & 79.5 & 80.0 & $+0.5$ \\
Llama-3.1-8B  & 77.2 & 81.3 & $+4.1$ \\
SmolLM3-3B    & 70.6 & 76.2 & $+5.6$ \\

\bottomrule
\end{tabular}%
}
\vspace{-0.1cm}
\caption{Hard without GT alignment (\%) under the standard 1.0-default and the recalibrated 0.5-default prompts, averaged across six judges per generator. $\Delta$ = (0.5-default) $-$ (1.0-default).}
\label{tab:c2-ceiling}
\vspace{-0.5cm}
\end{table}

\textbf{C2 ablation: 0.5-default without-GT prompt.} Table~\ref{tab:c2-ceiling} compares hard without-GT alignment under the standard $1.0$-default and a recalibrated $0.5$-default prompt. For the three strongest generators the delta is ${\leq}{+1.0}$\,pp, confirming structural task difficulty as the primary ceiling driver. For weaker generators ($+4.1$--$+5.6$\,pp), the $1.0$-default over-credits incorrect outputs; practitioners evaluating low-quality generators should consider the $0.5$-default prompt.

\textbf{RQ4: Does judge temperature affect alignment?}
\label{sec:abl:temperature}

\begin{figure}[h!]
\centering
\includegraphics[width=\columnwidth]{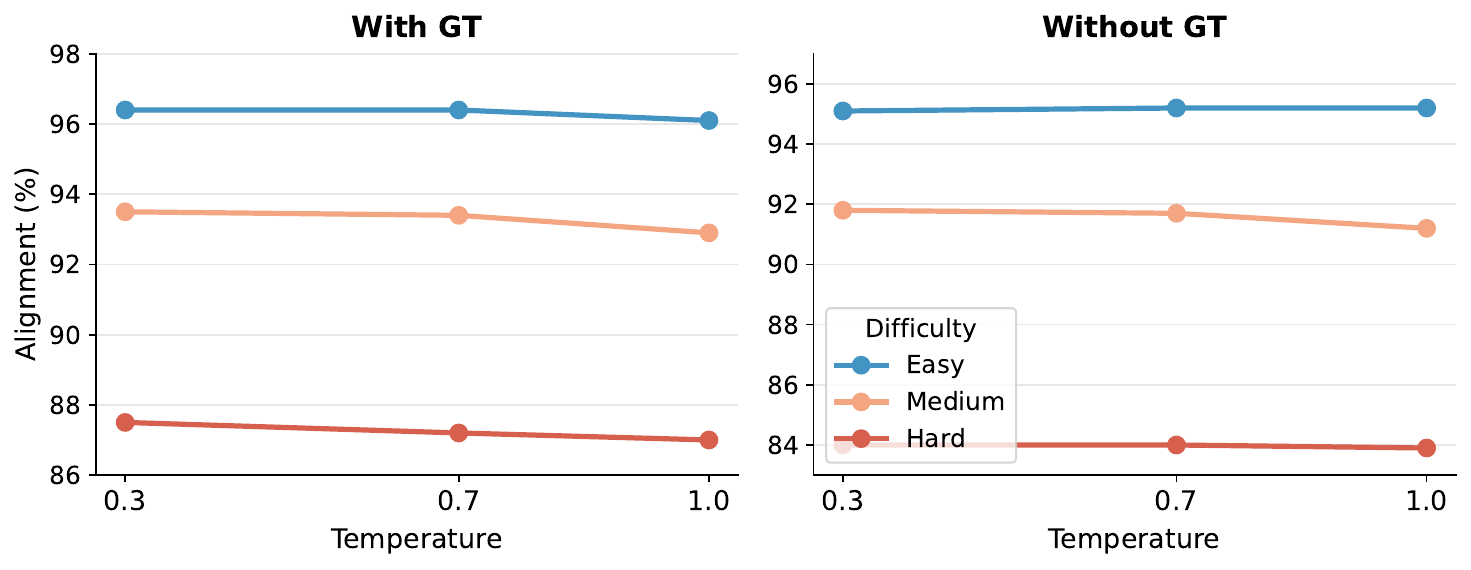}
\caption{%
\textbf{Judge temperature sensitivity.}
Qwen3-32B alignment at $T \in \{0.3, 0.7, 1.0\}$ on Llama-3.3-70B.
All three difficulty curves are near-flat; maximum spread $\leq 0.6$\,pp.}
\label{fig:abl-temperature}
\end{figure}

Figure~\ref{fig:abl-temperature} shows that Qwen3-32B alignment is insensitive to temperature across $T \in \{0.3, 0.7, 1.0\}$, with a maximum spread of $0.6$\,pp across all (difficulty, condition) cells; structural pattern-matching dominates (full table in Appendix~\ref{app:ablations}). A second (judge, generator) pairing (GPT-OSS-120B on Llama-3.1-8B-Instruct) confirms this, with a maximum spread of $0.25$\,pp across all three difficulty tiers, even tighter than the original pairing, indicating that temperature insensitivity is not an artefact of the single test-bed cell.

\begin{figure}[h!]
\centering
\includegraphics[width=\columnwidth]{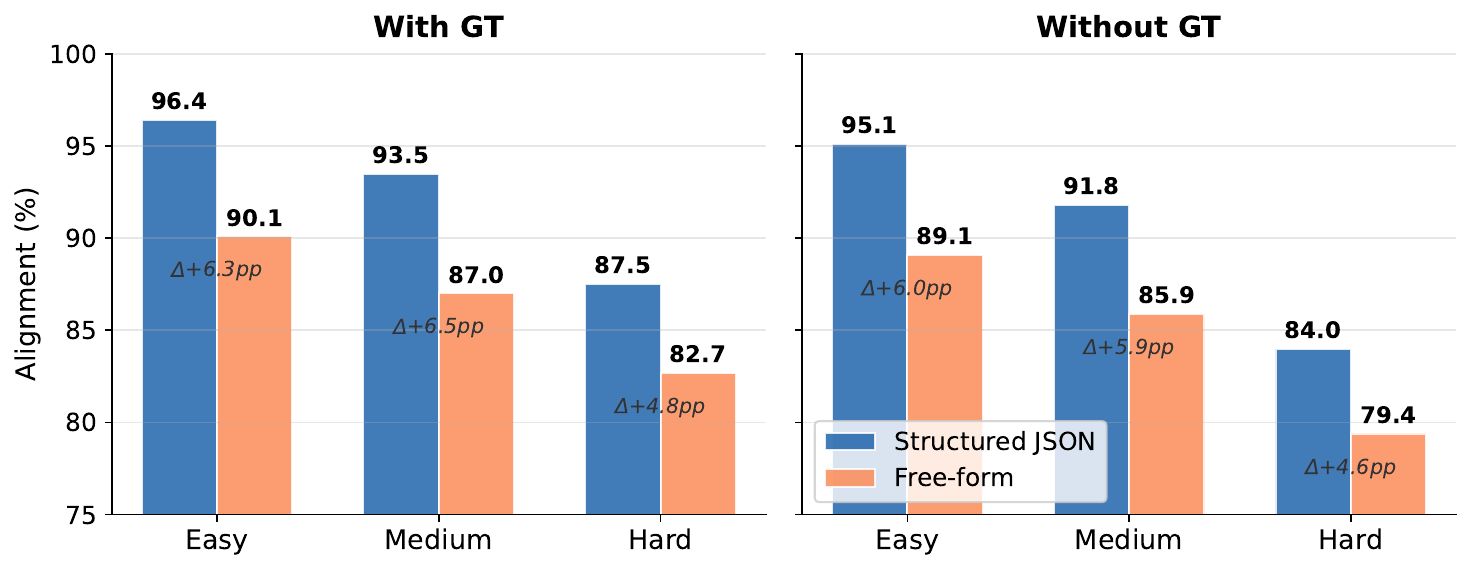}
\caption{%
\textbf{Prompt structure vs.\ alignment.}
Structured JSON prompt vs.\ free-form for Qwen3-32B on Llama-3.3-70B.
The per-metric rubric adds $+4.8$-$+6.5$\,pp with GT across all difficulty levels.}
\label{fig:abl-format}
\end{figure}

\textbf{RQ5: Does chain-of-thought reasoning in the judge improve alignment?}
\label{sec:abl:reasoning}

Table~\ref{tab:abl-reasoning} reports QwQ-32B alignment with thinking on vs.\ off across four open-weight generators and three difficulty levels (24 paired cells). CoT reasoning contributes negligibly, with a mean GT gap of $+0.11$\,pp and a maximum per-cell difference of $0.3$\,pp; no cell exceeds $0.3$\,pp on either condition. QwQ-32B's advantage in the main results therefore reflects training distribution rather than inference-time compute. The full per-generator grid is reproduced in Appendix~\ref{app:ablations}.

\begin{table}[t]
\centering
\small
\setlength{\tabcolsep}{3.5pt}
\renewcommand{\arraystretch}{1.02}

\resizebox{\columnwidth}{!}{%
\begin{tabular}{llcccccc}
\toprule

& &
\multicolumn{2}{c}{\textbf{Easy}} &
\multicolumn{2}{c}{\textbf{Medium}} &
\multicolumn{2}{c}{\textbf{Hard}} \\

\cmidrule(lr){3-4}
\cmidrule(lr){5-6}
\cmidrule(lr){7-8}

\textbf{Generator} &
\textbf{Setting} &
\textbf{GT} &
\textbf{No GT} &
\textbf{GT} &
\textbf{No GT} &
\textbf{GT} &
\textbf{No GT} \\

\midrule

\multirow{3}{*}{Llama-3.3-70B}
& Thinking on
  & 96.6 & 95.6 & 93.6 & 92.1 & 87.5 & 84.4 \\
& Thinking off
  & 96.5 & 95.6 & 93.6 & 92.0 & 87.3 & 84.4 \\
& $\Delta$
  & $+0.1$ & $0.0$ & $0.0$ & $+0.1$ & $+0.2$ & $0.0$ \\

\addlinespace[1.5pt]

\multirow{3}{*}{Llama-3.1-8B}
& Thinking on
  & 94.2 & 91.2 & 90.8 & 86.8 & 84.1 & 76.6 \\
& Thinking off
  & 94.3 & 91.4 & 90.8 & 86.8 & 83.9 & 76.7 \\
& $\Delta$
  & $-0.1$ & $-0.2$ & $0.0$ & $0.0$ & $+0.2$ & $-0.1$ \\

\addlinespace[1.5pt]

\multirow{3}{*}{Qwen3-32B}
& Thinking on
  & 95.0 & 93.9 & 90.9 & 89.1 & 84.7 & 81.1 \\
& Thinking off
  & 94.9 & 94.0 & 90.8 & 89.0 & 84.7 & 81.1 \\
& $\Delta$
  & $+0.1$ & $-0.1$ & $+0.1$ & $+0.1$ & $0.0$ & $0.0$ \\

\addlinespace[1.5pt]

\multirow{3}{*}{SmolLM3-3B}
& Thinking on
  & 90.0 & 86.7 & 84.8 & 80.3 & 77.2 & 69.7 \\
& Thinking off
  & 89.8 & 86.7 & 84.6 & 80.4 & 76.9 & 69.7 \\
& $\Delta$
  & $+0.2$ & $0.0$ & $+0.2$ & $-0.1$ & $+0.3$ & $0.0$ \\

\bottomrule
\end{tabular}%
}

\caption{Chain-of-thought reasoning in the QwQ-32B judge: alignment (\%) with thinking on vs.\ off across four open-weight generators. $\Delta$ = thinking-on $-$ thinking-off (pp).}
\label{tab:abl-reasoning}

\end{table}

\textbf{RQ6: Does prompt output format affect alignment?}
\label{sec:abl:format}

Figure~\ref{fig:abl-format} shows that the structured per-metric JSON prompt (verbatim in Appendix~\ref{app:prompt-gt}/\ref{app:prompt-without GT}) outperforms a free-form variant (Appendix~\ref{app:prompt-freeform}) by $+4.8$--$+6.5$\,pp with GT across all difficulty levels on the original (Qwen3-32B, Llama-3.3-70B) pairing, the largest lever we test among the configuration choices in this study. A second pairing (QwQ-32B on SmolLM3-3B) partially replicates this: structured format still wins on easy ($+3.9$\,pp) and medium ($+2.4$\,pp), but the effect shrinks relative to the first pairing and \emph{reverses} on hard ($-0.8$\,pp, free-form marginally ahead). We therefore do not treat prompt format as a uniformly dominant, difficulty-independent lever: the effect is real and judge/generator-dependent, largest on easier queries, and not guaranteed to generalise in direction on hard queries for every pairing. Full tables for both pairings are in Appendix~\ref{app:ablations}.
\section{Conclusion}
\label{sec:conclusion}
We introduce \textbf{AgentJudgeBench}, a benchmark measuring LLM-judge reliability on structured, dependency-driven tool-calling, where existing LLM-as-judge work offers little calibration, via three design choices: difficulty-stratified rewrites (easy/medium/hard, preserving the ground-truth trace), per-record DAG-topology annotation across six patterns, and a paired with-GT/without-GT evaluation protocol.
Across $321{,}648$ completed evaluations, four patterns emerge. First, alignment degrades monotonically with difficulty, approximately $1.5\times$ faster without ground truth than with it. Second, hard without-GT alignment converges to a $77$--$82\%$ band regardless of judge capacity, a task-level ceiling confirmed by the C2 recalibrated-prompt ablation (${\leq}{+1.0}$\,pp for capable generators). Third, GT exposure is counterproductive for frontier judges (Gemini-2.5-Pro: $-3.9$\,pp; GPT-5.4: $-1.5$\,pp), consistent with over-anchoring. Fourth, prompt structure is the largest configuration lever ($+4.8$--$+6.5$\,pp) but is judge/generator-dependent, while CoT reasoning and temperature are negligible (\S\ref{sec:exp:main}). \textbf{With ground truth, QwQ-32B agrees most closely with the programmatic reference and GPT-OSS-120B with human judgement; without it, frontier judges lead only narrowly.} Limitations and deployment guidance are in the Limitations section and Appendix~\ref{app:decision-guide}.
\section*{Supplementary Material}
\label{sec:release}

The Hugging Face dataset (link, page~1) contains all $3{,}808$ records, generator outputs, and all seven judges' with-GT/without-GT verdicts underlying every table and figure. The SyGra code release (link, page~1) contains the pipeline implementation -- data generation, difficulty rewriting, the programmatic judge (Eqs.~\ref{eq:prog-tool}-\ref{eq:alignment}), and the LLM-judge runner with full prompt templates (Appendix~\ref{app:prompts}) -- and a reproducibility script for the main results, under a permissive open-source license.
\section*{Limitations}
\label{app:limitations}

Four of five generators are open-weight; GPT-5.4 is a non-reproducible Azure snapshot (Appendix~\ref{app:judges}) that also serves as judge and rewrite meta-judge. All primary claims replicate on the open-weight generators independently of it; as a generator, GPT-5.4's self-bias when judged stays within range on other generators except sequence accuracy ($+0.172$ vs.\ $\leq+0.008$), a post-hoc correlational signal we cannot rule out and, being non-reproducible, cannot independently re-verify (Appendix~\ref{app:limitations-extended}).

We use a deterministic programmatic scorer, rather than a human or LLM annotator, as the reference signal, since only it scales to $321{,}648$ evaluations without reintroducing the reliability question under study. A 120-record single-annotator human study (Appendix~\ref{app:prog-judge-validation}) validates it at $92.5$--$98.3\%$ agreement on three of four metrics but only $82.5\%$ on parameter structure, where the scorer penalises schema-valid extra keys that annotators accept ($9$ of $21$ tool-selection disagreements reflect an analogous tool-redundancy gap); this does not drive our findings (ranking shifts $\leq0.09$\,pp under a maximally generous correction), though headline \emph{absolute} numbers still reflect the current scoring rule, so we report both programmatic and human-verdict rankings for best-judge claims. Extended discussion is in Appendix~\ref{app:limitations-extended}.

Records are synthetically generated rather than mined from enterprise traces, since a verified per-record reference -- required by our paired GT/no-GT protocol -- is not obtainable at this scale from proprietary systems; a two-level quality gate and the same human study ($92.7\%$ agreement) validate the pipeline, but domain drift remains an open concern (Appendix~\ref{app:limitations-extended}). Our six main judges are also all general-purpose: adding Prometheus-2~\citep{kim2024prometheus2} as a judge-specialised baseline shows it does not cluster with them ($20$--$30$\,pp lower alignment; Table~\ref{tab:interjudge}) and it is excluded from six-judge statistics (Appendix~\ref{app:limitations-extended}).

The RQ4/RQ6 ablations were each extended to a second (judge, generator) pairing: temperature insensitivity replicates tightly, but the prompt-format advantage only partially replicates and reverses on hard queries, so we treat format as judge/generator-dependent rather than uniformly dominant. Difficulty rewriting is unanimously validated on $58.1\%$ of easy$\to$medium and $93.9\%$ of medium$\to$hard records, so medium should be read as a robustness check rather than a fully calibrated tier. The without-GT ceiling is also partly prompt-dependent: a C2 ablation (Table~\ref{tab:c2-ceiling}) shifts it $\leq{+1.0}$\,pp under an alternative $0.5$-default prompt for the three strongest generators but $+4.1$--$+5.6$\,pp for the two weakest, so task difficulty is the primary ceiling driver only for capable generators (Appendix~\ref{app:limitations-extended}).

Finally, we study judge reliability at evaluation time only and do not test whether these failure modes carry over as a training signal (e.g., a reward model or model-selection gate); we reason through three findings under this framing -- GT-exposure over-anchoring, the without-GT verdict-compression ceiling, and correlated cross-judge failure -- and flag it as concrete future work in Appendix~\ref{app:limitations-extended}.

\section*{Ethical Considerations}

\textbf{Over-reliance on automated evaluation.}
Our findings reveal systematic failure modes in LLM judges, including the 77--82\% without-GT ceiling and over-anchoring in frontier models; practitioners unaware of these limitations risk certifying incorrect tool-calling outputs as correct, particularly in safety-critical domains (e.g., energy grid, healthcare) represented in our dataset. We mitigate this via deployment guidance (Appendix~\ref{app:decision-guide}) and by recommending against sole reliance on any single judge.

\textbf{Evaluation monoculture.} All six judges converge to similar failure patterns on hard queries, suggesting shared training-distribution biases; over-reliance on our rankings could reinforce a monoculture where the same blind spots propagate across pipelines, so we encourage complementing LLM judges with programmatic scoring and human review.

\textbf{Dual use, synthetic data, and compute.} The detailed failure-mode analysis (over-anchoring, the 1.0-default bias, prompt-format sensitivity) could be exploited to game judge-based evaluation; we release all prompts, scorer code, and raw outputs to enable countermeasures. All records are synthetically generated -- avoiding real-user privacy concerns but inheriting the generation pipeline's biases and possibly under-representing non-English or marginalised workflows. The full evaluation (321,648 paired judge calls) is compute-intensive; our decision guide (Appendix~\ref{app:decision-guide}) reduces unnecessary evaluation by recommending specific judges per scenario.

\section*{Acknowledgements}

We are especially grateful to Sai Rajeswar for consistently supporting research efforts like this one, and for generously taking the time to review the paper and help improve it whenever needed.

Generative AI tools (Claude) were used for language polishing and proofreading of author-written text; all research ideas, experiments, analysis, and writing are the authors' own.

\bibliographystyle{acl_natbib}
\bibliography{references}

\appendix

\section{Practitioner Decision Guide}
\label{app:decision-guide}

In most deployment scenarios ground-truth tool-call sequences are unavailable at inference time; the without GT judge is therefore the practical baseline. Table~\ref{tab:decision-guide} reports the best-performing judge for each deployment scenario, derived directly from Table~\ref{tab:full-results}. Rankings are computed as an \emph{unweighted} mean of per-cell alignment over all five generators per (condition, difficulty) combination; count-weighted averaging shifts results by $\leq 0.3$\,pp.

\begin{table}[!htbp]
\centering
\small
\setlength{\tabcolsep}{5pt}
\renewcommand{\arraystretch}{1.2}
\begin{tabularx}{\columnwidth}{@{}llXX@{}}
\toprule
\textbf{Condition} & \textbf{Difficulty} & \textbf{Best judge} & \textbf{Runner-up} \\
\midrule
Without GT & Easy   & Gemini-2.5-Pro (92.1\%) & GPT-5.4 (92.0\%)      \\
Without GT & Medium & GPT-5.4 (87.7\%)        & Gemini-2.5-Pro (87.4\%) \\
Without GT & Hard   & GPT-5.4 (79.7\%)        & Gemini-2.5-Pro (78.8\%) \\
\midrule
With GT    & Easy   & QwQ-32B (94.0\%)        & GPT-OSS-120B (93.0\%) \\
With GT    & Medium & QwQ-32B (90.0\%)        & GPT-OSS-120B (89.4\%) \\
With GT    & Hard   & QwQ-32B (83.4\%)        & GPT-OSS-120B (83.2\%) \\
\bottomrule
\end{tabularx}
\caption{Best judge per deployment scenario. The without GT rows reflect the common deployment setting where ground-truth tool calls are unavailable. GT = ground-truth tool calls available; without GT = judge uses only query and tool schemas. Alignment (\%) is unweighted mean over five generators.}
\label{tab:decision-guide}
\end{table}

\textbf{Takeaways.} (1)~\textbf{No ground truth} (the common case): Gemini-2.5-Pro and GPT-5.4 lead narrowly ($\leq 1$\,pp), but the without GT convergence ceiling (\S\ref{sec:exp:ceiling}) makes judge choice less consequential on hard queries: the practical difference between any two judges is $\leq 2$\,pp. (2)~\textbf{Ground truth available}: QwQ-32B dominates all difficulty levels; GPT-OSS-120B is the best open-weight alternative without reasoning-model inference overhead. (3)~\textbf{Single judge across all conditions}: QwQ-32B offers the best mean GT alignment (89.1\%) with competitive without GT performance (85.6\%). (4)~\textbf{Binary pass/fail pipelines}: Practitioners scoring verdicts on a binary pass/fail scale rather than $\{0, 0.5, 1\}$ should prefer GPT-OSS-20B over QwQ-32B. Under binary verdict collapse, judge rankings differ substantially from the three-point scale (Spearman $\rho = 0.03$; Table~\ref{tab:abl-binary}, Appendix~\ref{app:abl-binary}): judges with high 0.5-verdict rates (GPT-OSS-20B: $26.5\%$; GPT-5.4: $28.2\%$) gain a systematic advantage because every 0.5 verdict is remapped toward the programmatic direction, inflating their binary alignment. QwQ-32B's structural advantage over GPT-OSS-20B diminishes or reverses under binary scoring. (5)~\textbf{Human-intuition alignment}: Practitioners whose primary concern is agreement with human annotators rather than with the programmatic scorer should prefer GPT-OSS-120B, which tops both the programmatic and human-verdict orderings ($\Delta = -2.8$\,pp vs.\ human; Table~\ref{tab:judge-vs-human}, Appendix~\ref{app:prog-judge-validation}).

\section{Extended Limitations Discussion}
\label{app:limitations-extended}

This appendix expands, point by point, the summary given in the main-text Limitations section.

\textbf{Programmatic reference vs.\ human correctness.} We adopt a deterministic programmatic scorer, rather than a human or LLM annotator, as the reference signal, since it uniquely scales to $321{,}648$ evaluations without reintroducing the judge-reliability question under study. A 120-record human study (Appendix~\ref{app:prog-judge-validation}), each record scored by a single annotator, validates this choice at $92.5$--$98.3\%$ agreement on three of four metrics but only $82.5\%$ on parameter structure, where the scorer penalises schema-valid extra argument keys that annotators accept; the $92.7\%$ figure and human-verdict rankings (including the QwQ-32B 1st$\to$4th reversal) therefore reflect one calibrated annotator, not a consensus reference, and a multi-annotator replication is left to future work. This gap does not drive our findings: excluding parameter structure changes the aggregate ranking by $\leq0.2$\,pp (Table~\ref{tab:param-robustness}), and a maximally generous upper-bound correction moves every judge's alignment by $\leq0.09$\,pp with the ranking unchanged (Table~\ref{tab:eq2-bound})---though every headline \emph{absolute} number still incorporates the current, imperfectly-validated scoring rule, so we report both programmatic and human-verdict rankings wherever a best-judge claim is made.

\textbf{Tool redundancy.} The scorer also does not model tool redundancy: a generator reaching an equivalent result via a structurally different tool (e.g., a dedicated aggregation tool vs.\ composing two simpler ones) can be penalised by $p^{\text{tool}}$ and $p^{\text{seq}}$ even though the substitution is conceptually correct. The human study already surfaces a version of this ($9$ of $21$ tool-selection disagreements involve semantically equivalent tools or defensible extra calls; Appendix~\ref{app:prog-judge-validation}); quantifying corpus-wide prevalence and building a schema-aware tool-equivalence correction are left to future work.

\textbf{Synthetic data and real-trace validity.} Records are synthetically generated rather than mined from enterprise traces, since a verified reference for every record -- required by our paired GT/no-GT protocol -- is not obtainable at this scale from proprietary systems. A two-level quality gate and the same human study ($92.7\%$ agreement) validate the pipeline, but domain drift from real deployments remains an open validity concern, and a real-trace comparison is blocked on corpus access. All prompts are scoped to single-shot, stateless planning; headline findings are comparative rather than absolute-magnitude, which limits but does not eliminate exposure to this gap.

\textbf{Judge-specialised baseline.} Our six main judges are all general-purpose; to test generalisation to judge-specialised models we added Prometheus-2~\citep{kim2024prometheus2} across all five generators and both GT conditions (Table~\ref{tab:full-results}, row Prom2). It does not cluster with the general-purpose judges under either condition ($20$--$30$\,pp lower alignment, near-chance pairwise agreement; Table~\ref{tab:interjudge}), concentrated on parameter structure and query coverage (Table~\ref{tab:e4-permetric}), and shows a negative GT lift opposite to most judges -- consistent with, not contradicting, its own published reference-free result against \emph{human} judgement~\citep[Appendix F]{kim2024prometheus2}, a different comparison axis. It is excluded from all six-judge statistics and does not affect any headline finding; a broader judge-specialised roster is left to future work.

\textbf{GPT-5.4's triple role.} GPT-5.4 simultaneously serves as generator, judge, and rewrite meta-judge. Its self-bias on its own generations falls within its range on other generators (Appendix~\ref{app:self-preference}), except for sequence accuracy, where it over-credits itself ($+0.172$ vs.\ $\leq+0.008$ elsewhere); we find no aggregate self-preference but cannot rule out this metric-localised effect. This is a post-hoc correlational analysis of the existing grid, not a controlled ablation isolating GPT-5.4's role, and is further compounded by GPT-5.4 being a non-reproducible Azure snapshot with no fixed version string (Appendix~\ref{app:judges}): unlike other effects bounded in this section, this signal cannot be independently re-verified.

\textbf{Ablation coverage.} The RQ4 (temperature) and RQ6 (prompt format) ablations were each extended to a second (judge, generator) pairing: temperature insensitivity replicates more tightly, while the prompt-format advantage replicates only partially and reverses on hard queries, so we treat it as judge/generator-dependent rather than a uniformly dominant lever. A full grid across all judges and generators is left to future work.

\textbf{Difficulty-tier calibration.} Difficulty rewriting is unanimously validated on $58.1\%$ of easy$\to$medium and $93.9\%$ of medium$\to$hard records; the latter is our primary difficulty-degradation evidence, and the ``three difficulty tiers'' framing should be read with medium as a robustness check rather than a fully independent, calibrated tier.

\textbf{Without-GT ceiling and prompt default.} The without-GT ceiling is partly attributable to the prompt's default-to-$1.0$ rubric: the C2 ablation (Table~\ref{tab:c2-ceiling}) shows it shifts by only ${\leq}{+1.0}$\,pp under an alternative $0.5$-default for the three strongest generators, but by $+4.1$ and $+5.6$\,pp for the two weakest (Llama-3.1-8B, SmolLM3-3B); task difficulty remains the primary driver for capable generators, but the ceiling is somewhat prompt-dependent on weaker ones, a nuance the abstract's ``structural ceiling'' framing compresses. Headline figures use unweighted topology averages over an intentionally imbalanced corpus; count-weighted averaging shifts figures by $\leq1.5$\,pp with no ranking change (Appendix~\ref{app:pertopology-table}).

\textbf{Evaluation-time vs.\ training-time risk.} We study judge reliability at evaluation time and do not test whether these failure modes carry over when the same judges are used as a training signal (e.g., a reward model in RLHF/DPO fine-tuning of a tool-calling agent, or an automatic gate for model selection) -- an evaluation-time error changes a single reported number, while a training-time error changes the objective an agent is optimized against. Three findings would shift this risk differently: GT-exposure over-anchoring (Finding~2) would only affect pipelines that feed the reference trace into the reward model, pushing the agent toward mimicking surface form rather than penalising missing steps (case studies in Appendix~\ref{app:anchoring-cases}); the without-GT verdict-compression ceiling (\S\ref{sec:exp:ceiling}) is the higher-risk case, since a reward model built on such a judge would supply near-constant reward on hard queries regardless of actual correctness, weakening the gradient exactly where the agent most needs correction; and the correlated failure across judges (six-judge ensemble matching, not exceeding, the best individual judge; \S\ref{sec:exp:ceiling}) implies this risk would not be mitigated by ensembling our evaluated judges, since they fail on the same records for the same structural reason. Testing this directly would require training runs with judge-derived rewards under controlled bias conditions, outside this paper's scope; we flag it as a concrete, motivated direction for future work.

\section{Dataset Statistics}
\label{app:dataset-stats}

\begin{figure}[!htbp]
\centering
\includegraphics[width=\linewidth]{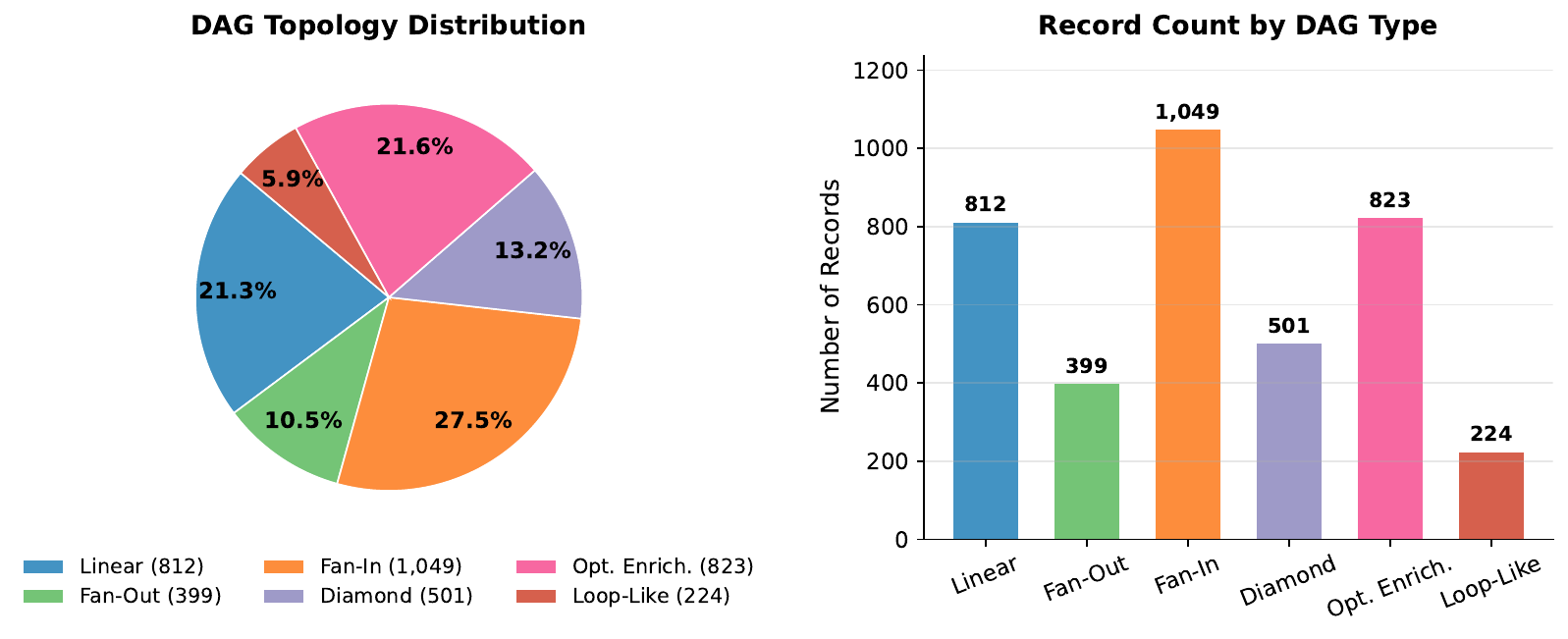}
\caption{DAG topology distribution across the 3,808 unique records (left: proportional pie, right: absolute counts). Fan-in accounts for over a quarter of all records; loop-like is the rarest at 5.9\%.}
\label{fig:dag-distribution}
\end{figure}

\begin{figure}[!htbp]
\centering
\includegraphics[width=\linewidth]{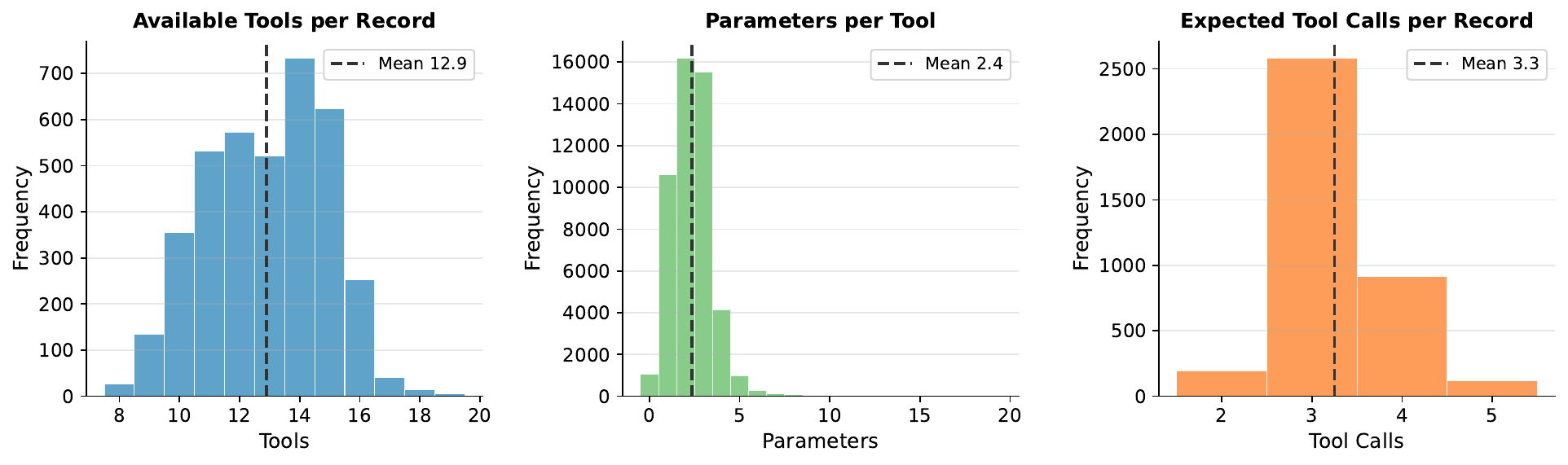}
\caption{Frequency distributions of (left) available tools per record, (centre) parameters per tool, and (right) expected tool calls per record. Dashed vertical lines mark the mean of each distribution. Tool inventory is tightly concentrated around 13 (range 8-19); parameter counts are right-skewed with a median of 2; expected call-sequence length ranges from 2 to 5.}
\label{fig:dataset-stats}
\end{figure}

This appendix reports the structural composition of the 3,808 AgentJudgeBench records, covering topology distribution, tool inventory size, parameter depth, and expected call-sequence length. All statistics are computed over the unique-record pool (i.e., collapsing the three difficulty rewrites of each record into one, since they share the same tool schemas and ground-truth trace). Figures~\ref{fig:dag-distribution} and~\ref{fig:dataset-stats} visualise the distributions; Tables~\ref{tab:dag-counts} and~\ref{tab:dataset-stats} report the precise counts.

\paragraph{Topology distribution (Table~\ref{tab:dag-counts}, Figure~\ref{fig:dag-distribution}).}

Records are not uniformly distributed across topologies. Fan-in (27.5\%) and optional enrichment (21.6\%) are the most prevalent, reflecting the frequency of multi-source aggregation and conditional enrichment patterns in the 15 enterprise seed domains. Loop-like records are the rarest (5.9\%) because iterative workflows are less commonly expressed as single-turn tool-call sequences in the generation pipeline. Linear records (21.3\%) serve as the structural baseline: they require no dependency tracking and are included to anchor the difficulty gradient. The imbalance is intentional: it mirrors the relative prevalence of each pattern in enterprise agentic workloads rather than imposing artificial uniformity.

\begin{table}[!htbp]
\centering
\small
\setlength{\tabcolsep}{8pt}
\renewcommand{\arraystretch}{1.2}
\begin{tabularx}{\columnwidth}{@{}Xrr@{}}
\toprule
\textbf{DAG Type} & \textbf{Records} & \textbf{\%} \\
\midrule
Linear              & \phantom{0}812  & 21.3 \\
Fan-Out             & \phantom{0}399  & 10.5 \\
Fan-In              & 1{,}049          & 27.5 \\
Diamond             & \phantom{0}501  & 13.2 \\
Optional Enrichment & \phantom{0}823  & 21.6 \\
Loop-Like           & \phantom{0}224  &  5.9 \\
\midrule
\textbf{Total}      & \textbf{3{,}808} & \textbf{100.0} \\
\bottomrule
\end{tabularx}
\caption{Record counts by DAG topology. Counts reflect unique records; each record has three difficulty variants (easy / medium / hard), giving $3{,}808 \times 3 = 11{,}424$ total rows in the full dataset.}
\label{tab:dag-counts}
\end{table}

\paragraph{Tool inventory and parameter depth (Table~\ref{tab:dataset-stats}, Figure~\ref{fig:dataset-stats}).}
Each record exposes a pool of 8-19 available tools (mean 12.9), of which the ground-truth trace invokes 2-5 (mean 3.3). Tools are deliberately over-provisioned: the ground truth uses on average only 25\% of the available pool, which forces the generator and judge to perform genuine tool selection rather than selecting by elimination. Each tool carries 0-19 typed parameters (mean 2.4, median 2); the small median reflects the prevalence of single-argument utility functions in the inventory, while the long tail (up to 19 parameters) comes from complex configuration and validation tools. The parameter minimum of zero corresponds to no-argument sentinel tools used in optional-enrichment and loop-like patterns. These structural properties collectively ensure that each of the four evaluation metrics (tool selection, parameter structure, sequence accuracy, and query coverage) is non-trivially exercised across the record pool.

\begin{table}[h!]
\centering
\small
\setlength{\tabcolsep}{8pt}
\renewcommand{\arraystretch}{1.2}
\begin{tabularx}{\columnwidth}{@{}Xrrrr@{}}
\toprule
\textbf{Quantity} & \textbf{Min} & \textbf{Max} & \textbf{Mean} & \textbf{Median} \\
\midrule
Available tools per record     &  8 & 19 & 12.9 & 13.0 \\
Parameters per tool            &  0 & 19 &  2.4 &  2.0 \\
Expected tool calls per record &  2 &  5 &  3.3 &  3.0 \\
\bottomrule
\end{tabularx}
\caption{Structural statistics of the 3,808 AgentJudgeBench records. All quantities are computed over unique records (collapsing difficulty variants). \emph{Available tools} is the size of the tool inventory exposed to the generator per record. \emph{Parameters per tool} counts typed argument fields in each tool's JSON schema. \emph{Expected tool calls} is the length of the ground-truth execution trace.}
\label{tab:dataset-stats}
\end{table}

\section{Evaluation Corpus Statistics}
\label{app:eval-stats}

\begin{table}[h!]
\centering
\small
\setlength{\tabcolsep}{8pt}
\renewcommand{\arraystretch}{1.2}
\resizebox{\columnwidth}{!}{%
\begin{tabular}{@{}lrrrrrr@{}}
\toprule
\textbf{Generator} & \textbf{Easy} & \textbf{Medium} & \textbf{Hard} & \textbf{Total} & \textbf{Max} & \textbf{Rate} \\
\midrule
Llama-3.3-70B  & 3{,}765 & 3{,}771 & 3{,}764 & 11{,}300 & 11{,}424 & 98.9\% \\
Llama-3.1-8B   & 3{,}707 & 3{,}698 & 3{,}606 & 11{,}011 & 11{,}424 & 96.4\% \\
Qwen3-32B      & 3{,}271 & 3{,}421 & 3{,}548 & 10{,}240 & 11{,}424 & 89.6\% \\
SmolLM3-3B     & 3{,}300 & 3{,}042 & 3{,}299 &  9{,}641 & 11{,}424 & 84.4\% \\
GPT-5.4        & 3{,}807 & 3{,}802 & 3{,}807 & 11{,}416 & 11{,}424 & 99.9\% \\
\midrule
\textbf{Total} & \textbf{17{,}850} & \textbf{17{,}734} & \textbf{18{,}024} & \textbf{53{,}608} & \textbf{57{,}120} & \textbf{93.8\%} \\
\bottomrule
\end{tabular}%
}
{\small\textit{Total: $53{,}608 \times 6\ \mathrm{judges} \times 2\ \mathrm{conditions} = 643{,}296$ evaluation instances (theoretical max); $321{,}648$ unique (generator, judge, difficulty, record) tuples, each evaluated under both with GT and without GT conditions.}}
\caption{Number of records with valid LLM-judge verdicts per generator and difficulty tier. \emph{Theoretical max} per generator is $3{,}808 \times 3 = 11{,}424$ (three difficulty variants of each base record). Judged records are identical across all six LLM judges for a given (generator, difficulty) cell. Multiplying the \emph{Total} column by $6$~judges~$\times~2$~conditions yields the full evaluation instance count per generator.}
\label{tab:eval-stats}

\end{table}

Table~\ref{tab:eval-stats} reports the exact number of records that received valid LLM-judge verdicts per (generator, difficulty) cell, after excluding records where the generator produced an unparseable or empty tool-call sequence. The base dataset contains $3{,}808$ unique records; each is rewritten into three difficulty variants, giving a theoretical maximum of $3{,}808 \times 3 = 11{,}424$ generator inputs per generator. The final column shows the success rate relative to this theoretical maximum.

\textbf{Sources of attrition.} The overall success rate is $93.8\%$ ($53{,}608$ of $57{,}120$). Attrition arises from two distinct sources at different pipeline stages:

\begin{itemize}
  \item \textbf{Unparseable generator output} (${\approx}5.7\%$ of inputs, weighted average): the generator produces a response that cannot be decoded as a valid JSON tool-call list. This is the dominant source of attrition, concentrated on SmolLM3-3B ($-15.6\%$) and Qwen3-32B ($-10.4\%$), and more prevalent at medium difficulty for SmolLM3-3B (medium rate $79.8\%$ vs.\ $86.6\%$ on easy/hard). These failures reflect intrinsic model capability gaps on the structured output format; re-generating these records would produce the same failure pattern and was not pursued.
  \item \textbf{Persistent LLM-judge null verdicts} (${\approx}3.3\%$ of main-grid judge calls): after the initial run, records with null \texttt{overall llm alignment percentage} were identified after retrying twice (two independent rerun rounds. Records that remained null after both retries (${\approx}10{,}498$ (generator, judge, difficulty, record) tuples in the main grid) were permanently excluded. These persistent nulls are concentrated on the longest GPT-5.4 and Llama-3.3-70B generator outputs, which push near the context limits of certain judge endpoints (particularly Claude Sonnet 4.5), causing consistent response truncation or malformed JSON. A third retry round was not run because (a) two retries had already demonstrated a ${<}5\%$ recovery rate for persistently null records, making further attempts cost-prohibitive, and (b) the persistent nulls are distributed uniformly across DAG topologies and difficulty levels, giving no reason to expect systematic bias in the excluded records.
\end{itemize}

\section{Bootstrap Confidence Intervals for Table~\ref{tab:full-results}}
\label{app:bootstrap-ci}

All alignment percentages in Table~\ref{tab:full-results} are means over $N_{g,d}$ per-record scores. To quantify uncertainty, we compute 95\% bootstrap confidence intervals ($n{=}2{,}000$ stratified resamples) for every (generator, judge, difficulty, condition) cell. Table~\ref{tab:ci-llama70b} reports the full CI matrix for the Llama-3.3-70B generator. The same bootstrap analysis applied to all five generators yields CIs $\leq 0.3$\,pp half-width throughout (the remaining matrices are in the supplementary code release), confirming that the point estimates in Table~\ref{tab:full-results} are reliable and that every monotone difficulty effect is statistically robust. CIs for the GT lift values (Eq.~\ref{eq:lift}) appear in Section~\ref{sec:exp:main}; all six lift CIs are non-overlapping across the positive-vs-negative divide.

\begin{table*}[h!]
\centering
\small
\resizebox{\textwidth}{!}{%
\setlength{\tabcolsep}{3.5pt}
\renewcommand{\arraystretch}{1.03}
\begin{tabular}{@{}lcccccc@{}}
\toprule
 & \multicolumn{2}{c}{\textbf{Easy}} & \multicolumn{2}{c}{\textbf{Medium}} & \multicolumn{2}{c}{\textbf{Hard}} \\
\cmidrule(lr){2-3}\cmidrule(lr){4-5}\cmidrule(lr){6-7}
\textbf{Judge} & \textbf{GT} & \textbf{No GT} & \textbf{GT} & \textbf{No GT} & \textbf{GT} & \textbf{No GT} \\
\midrule
GPT-5.4        & 91.4 [91.0,91.7] & 95.2 [94.8,95.6] & 87.4 [87.0,87.8] & 91.7 [91.2,92.2] & 83.0 [82.6,83.4] & 84.6 [83.9,85.2] \\
Claude Sonnet 4.5         & 93.3 [93.0,93.7] & 94.6 [94.3,95.0] & 89.9 [89.5,90.4] & 90.7 [90.2,91.2] & 83.7 [83.1,84.2] & 83.3 [82.6,83.9] \\
Gemini-2.5-Pro & 90.5 [90.0,91.0] & 95.7 [95.4,96.1] & 86.5 [86.0,87.0] & 92.2 [91.7,92.7] & 82.2 [81.6,82.7] & 84.6 [83.9,85.2] \\
QwQ-32B        & 96.6 [96.3,96.9] & 95.6 [95.2,95.9] & 93.6 [93.2,94.0] & 92.1 [91.6,92.5] & 87.5 [86.9,88.0] & 84.4 [83.7,85.1] \\
GPT-OSS-20B    & 89.0 [88.7,89.3] & 94.5 [94.1,94.9] & 87.2 [86.9,87.5] & 91.2 [90.7,91.7] & 83.9 [83.5,84.2] & 84.0 [83.3,84.7] \\
GPT-OSS-120B   & 96.0 [95.7,96.3] & 95.5 [95.2,95.9] & 93.0 [92.6,93.4] & 92.1 [91.7,92.6] & 87.2 [86.7,87.8] & 84.5 [83.8,85.2] \\
\bottomrule
\end{tabular}%
}
\caption{Bootstrap 95\% confidence intervals for alignment (\%) on the Llama-3.3-70B-Instruct generator. Format: mean [lo, hi].}
\label{tab:ci-llama70b}

\end{table*}

\section{Difficulty Degradation and Generator Accuracy}
\label{app:degradation}

\begin{figure}[!htbp]
\centering
\includegraphics[width=\linewidth]{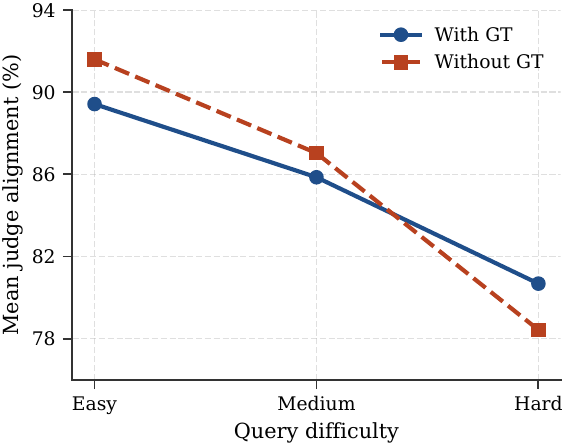}
\caption{Mean judge alignment as a function of query difficulty, averaged across all six judges and five generators. The without GT degradation slope is roughly $1.5\times$ steeper than the GT slope across all generators.}
\label{fig:degradation}
\end{figure}

Figure~\ref{fig:degradation} shows mean judge alignment as a function of query difficulty, averaged across all six judges and five generators. Table~\ref{tab:generator-accuracy} reports generator programmatic accuracy $\bar{p}$ (\%) on AgentJudgeBench, averaged across the four metrics and six DAG topologies.

\begin{table}[!htbp]
\centering
\small
\setlength{\tabcolsep}{4.5pt}
\renewcommand{\arraystretch}{1.05}
\begin{tabular}{@{}lrrr@{}}
\toprule
\textbf{Generator} & \textbf{Easy} & \textbf{Med.} & \textbf{Hard} \\
\midrule
Llama-3.3-70B & 96.1 & 92.5 & 84.5 \\
Qwen3-32B     & 93.7 & 88.5 & 80.0 \\
Llama-3.1-8B  & 89.3 & 85.2 & 74.4 \\
GPT-5.4       & 82.0 & 79.4 & 72.9 \\
SmolLM3-3B    & 83.5 & 77.2 & 65.5 \\
\bottomrule
\end{tabular}
\caption{Generator programmatic accuracy $\bar{p}$ (\%) on AgentJudgeBench, averaged across the four metrics and six DAG topologies. Generators are ranked by hard-difficulty accuracy.}
\label{tab:generator-accuracy}

\end{table}

\section{Programmatic Judge Validation}
\label{app:prog-judge-validation}

This appendix reports a human annotation study validating the programmatic scorer against independent human judgement. We sampled 120 \emph{hard}-difficulty records stratified across all six DAG topologies (20 per topology). For each record, annotators were shown the user query, available tool schemas, generated tool calls, ground-truth expected tool calls, and the programmatic scores. For each of the four metrics they were asked to \emph{agree} or \emph{disagree} with the programmatic score and provide a brief justification, yielding $120 \times 4 = 480$ metric-level verdicts. Records were stratified to over-represent cases where at least one programmatic score is below $1.0$ (14 per topology) alongside perfect-score records (6 per topology), ensuring annotators encountered the full range of difficulty. Annotators were not shown the paper's hypotheses or the LLM judge outputs prior to annotation. \textbf{Each record was scored by a single annotator; we did not collect a second, independent judgement per record and so cannot report an inter-annotator agreement statistic for the human labels themselves.} The $92.7\%$ and human-verdict-ranking figures throughout this appendix should be read as validating the programmatic scorer against one calibrated annotator's judgement, not against a consensus reference; a multi-annotator replication with an inter-annotator agreement check is left to future work (Limitations).

\begin{table}[h!]
\centering
\small
\setlength{\tabcolsep}{6pt}
\renewcommand{\arraystretch}{1.15}
\begin{tabularx}{\columnwidth}{@{}Xrrr@{}}
\toprule
\textbf{Metric} & \textbf{Agree} & \textbf{Disagree} & \textbf{Agreement \%} \\
\midrule
Tool Selection          & 111 &  9 & 92.5 \\
Parameter Structure     &  99 & 21 & 82.5 \\
Sequence Accuracy       & 117 &  3 & 97.5 \\
Query Coverage          & 118 &  2 & 98.3 \\
\midrule
\textbf{Overall}        & \textbf{445} & \textbf{35} & \textbf{92.7} \\
\bottomrule
\end{tabularx}

\begin{tabularx}{\columnwidth}{@{}Xrr@{}}
\toprule
\textbf{DAG Type} & \textbf{Agree / 80} & \textbf{\%} \\
\midrule
Linear              & 70 & 87.5 \\
Fan-Out             & 77 & 96.2 \\
Fan-In              & 75 & 93.8 \\
Diamond             & 77 & 96.2 \\
Optional Enrichment & 68 & 85.0 \\
Loop-Like           & 78 & 97.5 \\
\bottomrule
\end{tabularx}
\caption{Human annotator agreement with the programmatic scorer across 120 hard-difficulty records ($n{=}480$ metric-level verdicts), stratified at 20 records per DAG topology.}
\label{tab:prog-judge-agree}

\end{table}

The $92.7\%$ overall agreement across $480$ metric-level verdicts confirms the programmatic scorer as a reliable reference signal for the alignment metric of Section~\ref{sec:methodology:alignment}. Agreement is not uniform across metrics: \emph{parameter structure} is the weakest axis ($82.5\%$, $21$ disagreements), while \emph{sequence accuracy} and \emph{query coverage} are near-perfect ($97.5\%$ and $98.3\%$). Inspection of the $21$ parameter-structure disagreements reveals two error modes. The dominant one (${\approx}16$ of $21$ cases) is a \emph{schema-vs.-GT mismatch}: the model includes additional parameter keys that are valid per the available-tool JSON schema but absent from the specific ground-truth trace; the programmatic scorer penalises these extras (per Eq.~\ref{eq:prog-param-per-call}), whereas human annotators treat them as correct given the schema. The remaining $5$ disagreements involve structurally matching parameters that the scorer under-penalises due to partial-match rounding. This pattern is precisely the semantic-equivalence gap described in the Limitations section, and it bounds the scope of any mis-alignment attributable to scorer noise: $74.1\%$ of metric-level mismatches between the programmatic and LLM judges fall on tool selection, sequence accuracy, and coverage -- dimensions where structural and semantic correctness largely coincide.

On the tool-selection axis ($9$ disagreements), two failure modes emerge: semantic equivalence between tools with different names (e.g., \texttt{validate\_jurisdictional\_limits} vs.\ \texttt{assess\_jurisdictional\_limits}), and reasonable extra tools that the scorer penalises as out-of-reference. The per-DAG pattern is consistent with known topology difficulty: \emph{optional\_enrichment} has the lowest agreement ($85.0\%$), reflecting the inherent ambiguity in judging whether optional branches were correctly included or omitted, the same ambiguity that makes this topology moderately hard in the main judge evaluation (Table~\ref{tab:e5-pertopology}).

\paragraph{Judge alignment against human verdicts.}
The 120 annotated records also allow measuring how closely each LLM judge's with GT verdict matches human judgement, where records with annotator disagreement ($n{=}35$) are scored using the human-corrected value rather than the programmatic score. Table~\ref{tab:judge-vs-human} compares each judge's programmatic-reference alignment (primary metric) against their human-verdict alignment on the same 120 hard records.

\begin{table}[!htbp]
\centering
\small
\setlength{\tabcolsep}{5pt}
\renewcommand{\arraystretch}{1.05}
\begin{tabular}{@{}lrrr@{}}
\toprule
\textbf{Judge} & \textbf{Prog.} & \textbf{Human} & $\Delta$ \\
\midrule
GPT-5.4         & 78.7 & 71.2 & $-7.5$ \\
Claude Sonnet 4.5          & 78.1 & 74.5 & $-3.7$ \\
Gemini-2.5-Pro  & 78.3 & 68.0 & $-10.2$ \\
QwQ-32B         & 82.9 & 74.3 & $-8.6$ \\
GPT-OSS-20B     & 79.0 & 70.2 & $-8.8$ \\
GPT-OSS-120B    & 82.2 & 79.4 & $-2.8$ \\
\bottomrule
\end{tabular}
\caption{Judge alignment (\%, with GT) on 120 hard records: vs.\ programmatic reference and vs.\ human verdicts (35 corrected). $\Delta$ = human $-$ programmatic. Smaller $|\Delta|$ is better.}
\label{tab:judge-vs-human}
\end{table}

All six judges are systematically closer to the programmatic scorer than to human annotators (mean $\Delta = -6.9$\,pp). GPT-OSS-120B is the most human-aligned judge ($\Delta = -2.8$\,pp), while Gemini-2.5-Pro diverges most ($\Delta = -10.2$\,pp). The negative $\Delta$ reflects that human annotators are more lenient on hard examples (particularly on sequence accuracy and parameter structure), while both the programmatic scorer and LLM judges apply stricter structural matching. The judge rankings under human-verdict alignment differ meaningfully from programmatic-reference rankings (Spearman $\rho = 0.26$): QwQ-32B, which leads the programmatic leaderboard, drops to 4th under human alignment, while GPT-OSS-120B tops both orderings. This suggests that the programmatic scorer's strictness inflates QwQ's apparent advantage; GPT-OSS-120B is the most robust judge across both evaluation bases. Practitioners preferring alignment with human intuition should favour GPT-OSS-120B, regardless of whether programmatic or human verdicts are used as the reference.

\paragraph{Is the ranking driven by the parameter-structure gap?}
The result above raises a natural follow-up: since parameter structure is the one metric where the programmatic scorer measurably diverges from human judgement (\S\ref{sec:methodology:programmatic}, above), does QwQ-32B's programmatic-reference lead depend on how much weight that metric carries? We recompute the with-GT judge ranking (full $321{,}648$-evaluation grid, not just the 120-record human subset) under the equal-weight default, parameter structure downweighted by half, and parameter structure excluded entirely.

\begin{table}[h!]
\centering
\small
\setlength{\tabcolsep}{4pt}
\renewcommand{\arraystretch}{1.1}
\resizebox{\columnwidth}{!}{%
\begin{tabular}{@{}llc@{}}
\toprule
\textbf{Weighting} & \textbf{Full ranking (with GT)} & \textbf{Top judge} \\
\midrule
Equal (default)  & QwQ $>$ OSS-120B $>$ Claude $>$ GPT-5.4 $>$ OSS-20B $>$ Gemini & QwQ-32B (89.3\%) \\
Param $\times 0.5$ & QwQ $>$ OSS-120B $>$ Claude $>$ GPT-5.4 $>$ OSS-20B $>$ Gemini & QwQ-32B (89.2\%) \\
Param excluded   & QwQ $>$ OSS-120B $>$ Claude $>$ GPT-5.4 $>$ OSS-20B $>$ Gemini & QwQ-32B (89.1\%) \\
\bottomrule
\end{tabular}%
}
\caption{With-GT judge ranking (programmatic reference) under alternative weightings of the parameter-structure metric. The full six-judge ranking is identical across all three weightings.}
\label{tab:param-robustness}
\end{table}

The full six-judge ranking is unchanged under all three weightings, and QwQ-32B's score shifts by at most $0.2$\,pp between the equal-weight and parameter-excluded columns. The programmatic-reference ranking is therefore not an artefact of how parameter structure is weighted within the programmatic scorer. This is a different question from the one immediately above: reweighting the four metrics within the same (programmatic) reference does not change the ranking, but \emph{replacing} the reference with human verdicts on the 120-record subset does (QwQ-32B 1st$\to$4th). The two results together localize the issue precisely: the programmatic scorer's four metrics are not internally biasing the ranking against each other, but the programmatic scorer as a whole does disagree with human judgement on some records, and that disagreement is concentrated on parameter structure specifically.

\paragraph{Bounding the Eq.~2 rescoring fix.} A distinct, related question is what would happen if the parameter-structure scoring rule itself were corrected, crediting schema-valid extra argument keys as $1.0$ rather than the current $0.5$, rather than just reweighting the existing scores as above. A full fix requires checking each extra key against the tool's schema, which we leave to future work (main text, ``Programmatic scorer as reference''), but its impact is boundable now: we recompute the programmatic scorer's parameter-structure score under the maximally generous upper bound, crediting \emph{every} currently-penalised extra-key case as valid, not only the schema-valid subset a real fix would credit, across the full $53{,}619$-record, five-generator corpus. Only $0.83\%$ of all $186{,}654$ generated tool-calls exhibit an extra-key-only pattern at all, and the scorer's own mean parameter-structure accuracy moves by $+0.49$\,pp ($90.5\%\to91.0\%$). Table~\ref{tab:eq2-bound} propagates this upper bound into each judge's with-GT alignment, holding every judge's own verdicts fixed.

\begin{table}[h!]
\centering
\small
\setlength{\tabcolsep}{4pt}
\renewcommand{\arraystretch}{1.1}
\resizebox{\columnwidth}{!}{%
\begin{tabular}{@{}lccc@{}}
\toprule
\textbf{Judge} & \textbf{Current} & \textbf{Upper bound} & \textbf{$\Delta$} \\
\midrule
QwQ-32B         & 88.80 & 88.88 & $+0.079$ \\
GPT-OSS-120B    & 87.87 & 87.92 & $+0.046$ \\
Claude Sonnet 4.5 & 85.57 & 85.60 & $+0.030$ \\
GPT-5.4         & 84.83 & 84.87 & $+0.042$ \\
GPT-OSS-20B     & 84.11 & 84.20 & $+0.091$ \\
Gemini-2.5-Pro  & 81.95 & 81.99 & $+0.034$ \\
\bottomrule
\end{tabular}%
}
\caption{With-GT alignment (\%) under the current parameter-structure rule vs.\ the maximally generous upper-bound correction, record-weighted across the full corpus. Judges' own verdicts are held fixed; only the programmatic reference's parameter-structure score changes. Ranking order is identical to the current one in both columns.}
\label{tab:eq2-bound}
\end{table}

Every judge moves by at most $0.09$\,pp, and the ranking order (QwQ-32B $>$ GPT-OSS-120B $>$ Claude Sonnet 4.5 $>$ GPT-5.4 $>$ GPT-OSS-20B $>$ Gemini-2.5-Pro) is identical to the current one. Since this is the most generous correction possible (the true schema-aware fix can only credit a subset of what this upper bound credits), the real fix's effect is bounded above by these deltas. The Eq.~2 rescoring remains a committed future-work item, but the ranking-relevant risk it poses is now a measured sub-0.1-pp bound rather than an open question.

\section{Over-Anchoring Case Studies}
\label{app:anchoring-cases}

This appendix presents three representative records from the 1,322 over-anchoring instances identified on Gemini-2.5-Pro hard-difficulty outputs (records where $p^m_r < 1.0$, $\ell^m_{\text{GT}} > p^m_r$, and $\ell^m_{\text{without GT}} \leq p^m_r + 0.1$ on at least one metric). Cases 1--2 document the dominant \emph{sequence-accuracy anchor} mechanism; Case 3 documents the secondary \emph{coverage anchor}. In each case the judge's with-GT justification contains explicit reference to the GT trace (``sequence is logical given the reference''), whereas the without-GT justification independently identifies the structural flaw the programmatic judge also penalises. For each case, the \bad{red}-shaded box is the with-GT verdict (over-anchored) and the \good{green}-shaded box is the without-GT verdict (independently correct diagnosis); within each box, \bad{red} marks the reasoning that over-anchors to the reference, and \good{green} marks the reasoning that catches the actual structural flaw. GPT-5.4 shows a comparable negative GT lift overall ($-1.5$\,pp, Finding 2), but we have not extracted matching qualitative case studies for it here; the mechanism below is demonstrated for Gemini-2.5-Pro only, and its generalisation to GPT-5.4 is not independently verified.

\subsection*{Case 1: Sequence Anchor, Microbial-Growth Linear DAG}
\begin{querybox}[Query \& Generated Calls]
\footnotesize
\textbf{Query:} \emph{``Got this cheese with 250 CFU/g initial bugs, left at 20\textdegree C for 6.5 hours, need the growth rate, then a safe storage temperature for 120 hours, then the shelf life at that temp.''}

\textbf{Generated calls:} \texttt{calculate\_microbial\_growth\_rate} $\to$ \texttt{estimate\_optimal\_storage\_temperature} (2 calls; ground truth requires a third: \texttt{predict\_shelf\_life}).

\textbf{Programmatic score:} seq $= 0.67$ (partial match; missing third call).
\end{querybox}

\smallskip
\begin{wrongbox}[Gemini-2.5-Pro: With GT \hfill seq $=1.0$]
\footnotesize
\emph{``The sequence of the selected tools is logical, with the growth rate calculation correctly preceding the temperature estimation.''} The judge evaluates the two-call prefix against the GT prefix and deems it correct, \bad{ignoring the missing third step}.
\end{wrongbox}

\begin{correctbox}[Gemini-2.5-Pro: Without GT \hfill seq $=0.5$]
\footnotesize
\emph{``The plan includes a redundant, identical tool call, which is a minor structural flaw in the sequence.''} The judge \good{invents a different critique} rather than catching the missing call, but does not award full credit.
\end{correctbox}

\smallskip
\noindent\textit{Mechanism: GT exposure causes the judge to evaluate only the presented steps against the reference rather than checking for completeness.}

\subsection*{Case 2: Sequence Anchor, Cloud-Budget Linear DAG}
\begin{querybox}[Query \& Generated Calls]
\footnotesize
\textbf{Query:} \emph{``Cloud bill hit \$4,800 over 15 days vs.\ \$250/day normal. 10 days left, \$8,000 budget. What's going on and will we blow the budget?''}

\textbf{Generated calls:} \texttt{calculate\_budget\_deviation\_risk} (\texttt{predicted\_trend=1.93}) $\to$ \texttt{forecast\_budget\_usage}. The first call consumes a \texttt{predicted\_trend} argument that has not been computed by any prior step.

\textbf{Programmatic score:} seq $= 0.33$.
\end{querybox}

\smallskip
\begin{wrongbox}[Gemini-2.5-Pro: With GT \hfill seq $=1.0$]
\footnotesize
\emph{``The two selected tools are independent and can be executed in any order.''} The GT reference shows two calls and the judge rationalises independence; it does \bad{not notice the unresolved input dependency}.
\end{wrongbox}

\begin{correctbox}[Gemini-2.5-Pro: Without GT \hfill seq $=0.0$]
\footnotesize
\emph{``The first tool call depends on a \texttt{predicted\_trend} input that is not available from the query or generated by a prior step.''} Without the GT to anchor to, the judge \good{correctly identifies the data-dependency violation}.
\end{correctbox}

\smallskip
\noindent\textit{Mechanism: GT anchoring suppresses dependency-checking; the judge substitutes a plausibility heuristic (``independent tools can run in any order'') for structural verification.}

\subsection*{Case 3: Coverage Anchor, Power-Grid Linear DAG}
\begin{querybox}[Query \& Generated Calls]
\footnotesize
\textbf{Query:} \emph{``We're pulling 8,450 kW of 12,000 kW, expecting +1,200 kW, running hot for 5.3 h vs.\ 4 h threshold. Is everything okay?''}

\textbf{Generated calls:} Three analysis tools (load-duration risk, grid-fatigue risk, capacity headroom). Ground truth uses two; the extra call is \texttt{estimate\_grid\_fatigue\_risk}.

\textbf{Programmatic score:} cov $= 0.67$ (extra call penalised).
\end{querybox}

\smallskip
\begin{wrongbox}[Gemini-2.5-Pro: With GT \hfill cov $=1.0$]
\footnotesize
\emph{``The plan selects tools that address all numerical data points provided in the user's query.''} Seeing the GT calls, the judge \bad{credits the prediction for covering the query} rather than checking against the reference.
\end{wrongbox}

\begin{correctbox}[Gemini-2.5-Pro: Without GT \hfill cov $=0.5$]
\footnotesize
\emph{``The plan analyzes all data points but lacks a final synthesis tool to answer whether to be worried.''} Without GT, the judge applies an \good{independent completeness check} and finds partial coverage.
\end{correctbox}

\smallskip
\noindent\textit{Mechanism: GT presence triggers credit-by-association; the judge attributes coverage to the prediction by observing that the GT is complete, rather than evaluating the prediction independently.}

\section{Worked Example}
\label{app:worked-example}

This appendix walks through one complete record end to end -- query, available tools, generator output, ground truth, programmatic score, and LLM-judge verdict -- to make the abstract quantities in \S\ref{sec:methodology} concrete. Unlike the over-anchoring cases above, this record is \emph{not} a judge failure: the LLM judge and the programmatic scorer disagree because they resolve a genuine structural ambiguity differently, so the judge verdict below is shown in a neutral colour rather than as right or wrong.

\begin{querybox}[Query \hfill \textit{Hard, Fan-In DAG}]
\footnotesize
\emph{``Hey, so I bought something for \$184.73 yesterday morning around 9:45 UTC on the 18th of September 2023, and I'm wondering if that's weird for me since I usually do like 5 transactions a day and spend around \$312.50 daily on average, can you tell me how abnormal this looks overall?''}
\end{querybox}

\smallskip
\begin{querybox}[Available Tools \hfill \textit{Excerpt, 14 total}]
\footnotesize
\texttt{calculate\_temporal\_transaction\_deviation}, \texttt{calculate\_amount\_based\_deviation}, \texttt{compute\_composite\_behavior\_anomaly\_score}, plus 11 distractor tools (e.g., \texttt{detect\_device\_or\_location\_shift}, \texttt{get\_user\_behavior\_context}) not relevant to this query.
\end{querybox}

\smallskip
\begin{gtbox}[Ground-Truth Tool-Call Sequence]
\footnotesize
(1)~\texttt{calculate\_temporal\_transaction\_deviation}, (2)~\texttt{calculate\_amount\_based\_deviation}, (3)~\texttt{compute\_composite\_behavior\_anomaly\_score}: the first two calls are independent and must both complete before the third (fan-in), which combines their outputs.
\end{gtbox}

\smallskip
\begin{querybox}[Generator Output: Llama-3.3-70B-Instruct]
\footnotesize
The same three tools, in the order (1)~\texttt{calculate\_amount\_based\_deviation}, (2)~\texttt{calculate\_temporal\_transaction\_deviation}, (3)~\texttt{compute\_composite\_behavior\_anomaly\_score}: the first two calls are swapped relative to the reference, and the third call is emitted immediately with symbolic placeholder arguments (\texttt{\{\% output of function with id 2 \%\}}) rather than waiting for the first two to resolve.
\end{querybox}

\smallskip
\begin{gtbox}[Programmatic Judge \hfill tool $=1.0$, param $=1.0$, seq $=0.33$, cov $=1.0$]
\footnotesize
tool $=1.0$ (correct set, no extras); param $=1.0$ (all required argument names present -- placeholder values do not count against this metric by design, \S\ref{sec:methodology:programmatic}); \textbf{seq $=0.33$} (position-by-position match against the 3-call reference: only position 3 agrees); cov $=1.0$ (both source aspects of the query -- amount and timing -- are addressed).
\end{gtbox}

\smallskip
\begin{neutralbox}[LLM Judge: QwQ-32B, With GT \hfill tool $=1.0$, param $=1.0$, seq $=1.0$, cov $=1.0$]
\footnotesize
\emph{``Logical order: calculate individual deviations first, then combine them into a composite score.''} The without-GT verdict reaches the identical scores with an equivalent justification.
\end{neutralbox}

\smallskip
\noindent\textbf{Reading the divergence.} The LLM judge and the programmatic scorer agree on three of four metrics and disagree sharply on sequence accuracy (match score $0.33$, the sole source of this record's $83.3\%$ alignment rather than $100\%$). Both readings are defensible under different notions of ``correct sequencing'': the judge treats the plan as a coherent two-then-one dependency structure regardless of which independent call comes first, while the programmatic scorer enforces the exact reference ordering position by position. This is a concrete instance of the structural-vs-semantic gap discussed in Limitations: the disagreement is not a scorer bug, but a genuine ambiguity in how strictly ``sequence accuracy'' should be defined for fan-in topologies where sibling branches are interchangeable.

\section{C3 Corrupted-GT Control: Full Results}
\label{app:c3-table}

Table~\ref{tab:c3-anchoring} reports per-difficulty alignment for both judges under standard GT, without GT, and the corrupted-GT (C3) condition on the Llama-3.3-70B generator. The C3 condition replaces the judge's reference with a randomly sampled GT from a different record of the same DAG topology. The key finding is the judge-level split: Gemini-2.5-Pro with corrupted GT matches its standard GT alignment within $0$-$1.8$\,pp across all difficulties, while QwQ-32B with corrupted GT matches its without GT alignment within $0.2$\,pp. This confirms the mechanism described in Section~\ref{sec:exp:main}: Gemini anchors to any reference block regardless of content; QwQ exercises independent reasoning when the reference is incoherent.

\section{Notation}
\label{app:notation}

Table~\ref{tab:notation} summarises the symbols used throughout the paper, in the order in which they appear.

\begin{table*}[!htbp]
\centering
\small
\setlength{\tabcolsep}{6pt}
\renewcommand{\arraystretch}{1.25}
\resizebox{0.8\textwidth}{!}{%
\begin{tabular}{p{2cm}p{10cm}}
\toprule
\textbf{Symbol} & \textbf{Definition} \\
\midrule
$\mathcal{G}$
  & Set of generator models whose tool-call sequences are judged; $|\mathcal{G}| = 5$. \\
$\mathcal{J}$
  & Set of LLM judges evaluated; $|\mathcal{J}| = 6$. \\
$\mathcal{D}$
  & Difficulty levels of the query rewrites, $\mathcal{D} = \{\text{easy}, \text{medium}, \text{hard}\}$. \\
$g,\ j,\ d$
  & Individual generator, judge, and difficulty, with $g \in \mathcal{G}$, $j \in \mathcal{J}$, $d \in \mathcal{D}$. \\
$c$
  & Judge condition, $c \in \{\text{GT}, \text{without GT}\}$: whether the prompt includes the ground-truth tool calls. \\
$r$
  & Individual record (one query, its tool schemas, ground-truth and predicted tool calls). \\
$M$
  & Number of evaluation metrics; $M = 4$ (tool selection, parameter structure, sequence accuracy, query coverage). \\
$N_{g,d}$
  & Record count in the $(g, d)$ configuration after filtering unparseable generator outputs. \\
$p_r \in [0,1]$
  & Deterministic programmatic scorer's verdict on record $r$ (reference signal for alignment). \\
$\ell_{j,r,c} \in \{0, 0.5, 1\}$
  & LLM judge $j$'s verdict on record $r$ under condition $c$. \\
$\mathrm{align}(j, g, d, c)$
  & Mean per-record alignment percentage (Eq.~\ref{eq:alignment}). \\
$\overline{\mathrm{align}}_{c}(j)$
  & Alignment of judge $j$ under condition $c$, averaged over all $(g, d)$ configurations. \\
$\mathrm{lift}(j)$
  & Ground-truth lift of judge $j$: $\overline{\mathrm{align}}_{\text{GT}}(j) - \overline{\mathrm{align}}_{\text{without GT}}(j)$ (Eq.~\ref{eq:lift}). \\
$\Delta$(hard$-$easy)
  & Difficulty-induced degradation: $\mathrm{align}(\cdot,\cdot,\text{hard},\cdot) - \mathrm{align}(\cdot,\cdot,\text{easy},\cdot)$, in percentage points (pp). \\
\bottomrule
\end{tabular}%
}
\caption{Notation used in the paper.}
\label{tab:notation}
\end{table*}

\section{Judge LLMs Comparison}
\label{app:judge-comparison}

Table~\ref{tab:comparison} compares AgentJudgeBench against the seven existing
systems that deploy LLM judges for agentic tool-calling evaluation, across four
dimensions: per-metric decomposition, difficulty variation, ground-truth
ablation, and use of a deterministic programmatic reference. No prior system
addresses more than one of these dimensions; AgentJudgeBench is the first to
provide all four.

\begin{table*}[h!]
\centering
\footnotesize
\renewcommand{\arraystretch}{1.0}
\setlength{\tabcolsep}{4pt}
\resizebox{\textwidth}{!}{%
\begin{tabular}{l l c c c c}
\toprule
\textbf{System} & \textbf{Judge Model} & \textbf{Per-Metric} & \textbf{Difficulty} & \textbf{GT Abl.} & \textbf{Prog.\ Ref.} \\
\midrule
ToolEval \citep{qin2024toolllm} & ChatGPT & \xmark & \xmark & \xmark & \xmark \\
StableToolBench \citep{guo2024stabletoolbench} & GPT-4-turbo & \xmark & \xmark & \xmark & \xmark \\
MCP-AgentBench \citep{guo2025mcpagentbench} & LLM + Rules & \xmark & \xmark & \xmark & \xmark \\
GeoBenchX \citep{krechetova2025geobenchx} & 3-Judge Panel & \xmark & \xmark & \xmark & \xmark \\
Agent-as-a-Judge \citep{zhuge2024agentjudge} & Agent & \xmark & \xmark & \xmark & \xmark \\
Auto-Eval Judge \citep{bhonsle2025autoeval} & GPT-4o & \xmark & \xmark & \xmark & \xmark \\
ToolSandbox \citep{lu2024toolsandbox} & \textit{Replaced} & N/A & \xmark & N/A & \cmark \\
\midrule
\textbf{AgentJudgeBench (Ours)} & \textbf{Multi-scale LLMs (3B $\rightarrow$ Frontier)} & \cmark & \cmark & \cmark & \cmark \\
\bottomrule
\end{tabular}%
}
\caption{
Comparison of systems that deploy LLM judges for agentic tool-calling evaluation. \textbf{Judge Model}: LLM used as judge. \textbf{Per-Metric}: whether evaluation is decomposed into fine-grained dimensions. \textbf{Difficulty}: whether tasks span multiple difficulty tiers. \textbf{GT Abl.}: whether the effect of ground-truth availability is studied. \textbf{Prog.\ Ref.}: whether a deterministic evaluator is used as a bias-free baseline.
}
\label{tab:comparison}

\end{table*}

\section{Generator Models}
\label{app:generators}

Table~\ref{tab:generators} lists the five generator models used to produce tool-calling outputs. Models span four capability tiers -- small open-source (3B), mid-scale open (8B), large open (32B--70B), and frontier closed (GPT-5.4) -- ensuring the judge evaluation covers a representative range of output quality. All generators are decoded at temperature $0$ with the model's native function-calling prompt.

\begin{table*}[!htbp]
\centering
\small
\renewcommand{\arraystretch}{1.1}
\setlength{\tabcolsep}{5pt}
\resizebox{0.9\textwidth}{!}{%
\begin{tabular}{@{}lll@{}}
\toprule
\textbf{Short name} & \textbf{HF / Server identifier} & \textbf{Size} \\
\midrule
Llama-3.3-70B-Instruct & \texttt{meta-llama/Llama-3.3-70B-Instruct} & 70B \\
Qwen3-32B              & \texttt{Qwen/Qwen3-32B}                   & 32B \\
Llama-3.1-8B-Instruct  & \texttt{meta-llama/Llama-3.1-8B-Instruct} & 8B  \\
SmolLM3-3B             & \texttt{HuggingFaceTB/SmolLM3-3B}         & 3B  \\
GPT-5.4$^{\dagger}$    & Azure OpenAI / \texttt{gpt-5.4}, api-ver \texttt{2025-04-01-preview}, accessed Apr 2026 & - \\
\bottomrule
\end{tabular}%
}
\begin{flushleft}
\footnotesize $^{\dagger}$``GPT-5.4'' is the internal designation for the GPT-5 preview variant deployed on Azure OpenAI as of April 2026; not an officially published model version string. Raw per-record outputs are included in the supplementary data release.
\end{flushleft}
\caption{Generator models used to produce tool-calling outputs. All generators use temperature $=0$ with their native function-calling prompt.}
\label{tab:generators}

\end{table*}

\section{Rewriting-Preservation Validation Study}
\label{app:rewrite-validation}

This appendix reports the full results of the meta-judge validation study summarised in \S\ref{sec:methodology:data}. \emph{Goal:} verify, independently of the programmatic judge, that our difficulty-controlled rewrites (a)~preserve the ground-truth tool-call sequence and (b)~are strictly harder (less explicit) than their predecessor.

\begin{table*}[h!]
\centering
\small
\setlength{\tabcolsep}{6pt}
\renewcommand{\arraystretch}{1.1}
\resizebox{0.9\textwidth}{!}{%
\begin{tabular}{@{}lccc@{}}
\toprule
\textbf{Criterion} & \textbf{Claude Sonnet 4.5} & \textbf{GPT-5.4} & \textbf{Gemini-2.5-Pro} \\
\midrule
Medium preserves ground-truth tool calls        & 100.0\% & 94.9\% & 100.0\% \\
Hard preserves ground-truth tool calls          & 100.0\% & 77.8\% & 97.5\% \\
Medium strictly harder than easy                & 99.5\%  & 62.1\% & 84.8\%  \\
Hard strictly harder than medium                & 100.0\% & 97.5\% & 95.5\%  \\
\bottomrule
\end{tabular}%
}
\caption{Per-judge ``yes''-rate on the rewrite-validation criteria over $n = 198$ stratified triplets.}
\label{tab:rewrite-validation-per-judge}
\end{table*}

\begin{table*}[h!]
\centering
\small
\setlength{\tabcolsep}{6pt}
\renewcommand{\arraystretch}{1.1}
\resizebox{0.9\textwidth}{!}{%
\begin{tabular}{@{}lcccc@{}}
\toprule
\textbf{Criterion} & \textbf{All-yes} & \textbf{All-no} & \textbf{Mixed} & \textbf{Unanimity\%} \\
\midrule
Medium preserves ground-truth tool calls        & 188 & 0 & 10 & 94.9\% \\
Hard preserves ground-truth tool calls          & 152 & 0 & 46 & 76.8\% \\
Medium strictly harder than easy                & 114 & 1 & 83 & 58.1\% \\
Hard strictly harder than medium                & 186 & 0 & 12 & 93.9\% \\
\bottomrule
\end{tabular}%
}
\caption{Unanimity of the three meta-judges on each criterion ($n = 198$).}
\label{tab:rewrite-validation-unanimity}
\end{table*}
\paragraph{Protocol.} We stratify-sampled $198$ triplets ($33$ per DAG topology) from the $3{,}808$ record pool and asked each of three frontier LLM meta-judges (Claude Sonnet 4.5, GPT-5.4, Gemini-2.5-Pro) to emit a single JSON verdict per triplet, answering four Boolean questions: (i)~does the medium query admit the same ground-truth tool calls as the easy query?; (ii)~same for hard?; (iii)~is the medium rewrite strictly harder than easy (less explicit in at least one of parameter names, numeric values, or tool intents)?; (iv)~is hard strictly harder than medium? All three meta-judges receive identical prompt scaffolding and decoding parameters. The meta-judges are \emph{not} shown any generator's tool-call prediction; they reason only over the three query variants, the shared available-tools schema, and the shared ground-truth tool-call sequence.

\paragraph{Per-judge verdict rates.}
Table~\ref{tab:rewrite-validation-per-judge} reports the fraction of triplets for which each meta-judge answered ``yes'' to each question. Task-preservation rates are uniformly high ($77.8\%$-$100\%$ across all judges and both rewrite pairs); the strict-hardening rates are also high for hard-versus-medium ($\geq 95.5\%$) but lower and more judge-dependent for medium-versus-easy, driven by GPT-5.4's stricter interpretation of what constitutes a genuine loss of explicitness.

\paragraph{Unanimous agreement.}
Table~\ref{tab:rewrite-validation-unanimity} reports, per criterion, the number of triplets on which the three meta-judges unanimously agreed (all-yes or all-no) versus gave mixed verdicts. Crucially, \emph{no triplet receives a unanimous ``no'' verdict} on either task-preservation question, and every single triplet is unanimously judged as hard-strictly-harder-than-medium. The only criterion with sub-$60\%$ unanimity is medium-strictly-harder-than-easy, consistent with the per-judge analysis above: annotators disagree at the margin on whether medium constitutes a genuine hardening as opposed to a paraphrase.

\begin{table*}[!htbp]
\centering
\small
\renewcommand{\arraystretch}{1.1}
\setlength{\tabcolsep}{5pt}
\resizebox{0.9\textwidth}{!}{%
\begin{tabular}{@{}lll@{}}
\toprule
\textbf{Short name} & \textbf{Provider / Identifier} & \textbf{Size} \\
\midrule
GPT-5.4$^{\dagger}$ & Azure OpenAI / \texttt{gpt-5.4}, api-ver \texttt{2025-04-01-preview}, accessed Apr 2026 \\
Claude Sonnet 4.5  & Anthropic proxy / \texttt{claude-sonnet-4-5-20250929-v1:0}, accessed Apr 2026         \\
Gemini-2.5-Pro     & Vertex AI proxy / \texttt{gemini-2.5-pro-preview-05-06}, accessed Apr 2026            \\
QwQ-32B            & vLLM / \texttt{Qwen/QwQ-32B}, \texttt{enable\_thinking=true}    & 32B                     \\
GPT-OSS-20B        & vLLM / \texttt{openai/gpt-oss-20b}    & 20B                                               \\
GPT-OSS-120B       & vLLM / \texttt{openai/gpt-oss-120b}     & 120B                                             \\
\bottomrule
\end{tabular}%
}
\begin{flushleft}
\footnotesize $^{\dagger}$``GPT-5.4'' is the internal designation for the GPT-5 preview variant deployed on Azure OpenAI as of April 2026; it is not an officially published model version string.
\end{flushleft}
\caption{LLM judge configurations evaluated in this paper. All judges consume the same prompt scaffold (Appendix~\ref{app:prompts}) and produce the same JSON output schema.}
\label{tab:judges}

\end{table*}

\paragraph{Interpretation.}
The validation supports the stronger of our two design claims (task preservation) and partially supports the weaker one (strict hardening). The task-preservation result means that any alignment degradation observed in Section~\ref{sec:experiments} between easy and hard difficulty conditions is \emph{not} attributable to the rewrites silently changing the underlying task: even the most conservative meta-judge (GPT-5.4, which tends to reject rewrites for minor interpretative drift) finds task-preservation intact on $77.8\%$ of hard rewrites, and the unanimous-no rate is zero. The strict-hardening result for medium-vs-easy is a weaker claim; on that axis our rewriting process achieves a hardening GPT-5.4 would accept only about two-thirds of the time, suggesting that a subset of our medium rewrites are better characterised as paraphrases than as genuine hardenings. This is an informative finding on its own: it suggests the paper's ``easy $\rightarrow$ medium $\rightarrow$ hard'' degradation curves may partially conflate paraphrase-robustness with genuine difficulty-robustness, and is a natural target for a follow-up rewriting-pipeline revision.

\paragraph{Robustness to dropping the non-reproducible meta-judge.}
GPT-5.4 is a non-reproducible snapshot (Limitations) and also serves as a generator and LLM judge elsewhere in the pipeline, raising the question of whether the rewrite-validation rates above are dependent on it. We recompute unanimity using only the two reproducible meta-judges, Claude Sonnet 4.5 and Gemini-2.5-Pro.

\begin{table}[h!]
\centering
\small
\setlength{\tabcolsep}{4pt}
\renewcommand{\arraystretch}{1.1}
\resizebox{\columnwidth}{!}{%
\begin{tabular}{@{}lccc@{}}
\toprule
\textbf{Criterion} & \textbf{2-judge (no GPT-5.4)} & \textbf{3-judge (published)} & \textbf{$\Delta$} \\
\midrule
Medium preserves tool calls    & $100.0\%$ & $94.9\%$ & $+5.1$\,pp \\
Hard preserves tool calls      & $97.5\%$  & $76.8\%$ & $+20.7$\,pp \\
Medium harder than easy        & $84.8\%$  & $58.1\%$ & $+27.3$\,pp \\
Hard harder than medium        & $95.5\%$  & $93.9\%$ & $+1.5$\,pp \\
\bottomrule
\end{tabular}%
}
\caption{Unanimous-agreement rate on each rewrite-validation criterion, Claude Sonnet 4.5 + Gemini-2.5-Pro only ($n=198$), versus the published three-meta-judge rate. $\Delta = $ 2-judge $-$ 3-judge.}
\label{tab:rewrite-validation-no-gpt54}
\end{table}

All four deltas are positive: dropping GPT-5.4 \emph{raises} the unanimous-agreement rate on every criterion, most sharply on hard-preserves-tool-calls ($+20.7$\,pp) and medium-harder-than-easy ($+27.3$\,pp). This is the opposite of what a self-serving meta-judge would produce: GPT-5.4 is consistently the \emph{most conservative} of the three meta-judges (Table~\ref{tab:rewrite-validation-per-judge}), not one inflating agreement to validate its own downstream role. The published three-judge rates are therefore a lower bound driven by GPT-5.4's stricter interpretation, not evidence that the difficulty design is unreliable; the two independently-reproducible meta-judges alone would support a substantially stronger validation claim. We report the more conservative three-judge figures throughout the main text.

\section{GPT-5.4 Self-Preference Check}
\label{app:self-preference}

GPT-5.4 also acts as both a generator and an LLM judge in the main grid, raising a second, distinct concern from the meta-judge robustness check above: does GPT-5.4-as-judge over-credit GPT-5.4-as-generator? We test this directly using the existing $321{,}648$-evaluation grid (no new inference). For each (generator, metric) pair we compute the signed bias $\mathrm{llm} - \mathrm{prog}$ averaged over all records and difficulty tiers, comparing GPT-5.4-as-judge's bias on its own generations against its bias on the four other generators, with GPT-OSS-120B as a generator-agnostic control judge.

\begin{table}[h!]
\centering
\small
\setlength{\tabcolsep}{4pt}
\renewcommand{\arraystretch}{1.1}
\resizebox{\columnwidth}{!}{%
\begin{tabular}{@{}lrrrrr@{}}
\toprule
\textbf{Generator} & \textbf{tool} & \textbf{param} & \textbf{seq} & \textbf{cov} & \textbf{mean} \\
\midrule
Llama-3.3-70B    & $-0.080$ & $+0.052$ & $-0.066$ & $-0.001$ & $-0.024$ \\
Llama-3.1-8B     & $-0.034$ & $+0.070$ & $-0.060$ & $+0.009$ & $-0.004$ \\
Qwen3-32B        & $-0.034$ & $+0.056$ & $+0.008$ & $-0.018$ & $+0.003$ \\
SmolLM3-3B       & $+0.040$ & $+0.096$ & $-0.004$ & $+0.005$ & $+0.034$ \\
GPT-5.4$^{\ast}$ & $-0.026$ & $-0.019$ & $+0.172$ & $-0.077$ & $+0.012$ \\
\bottomrule
\end{tabular}%
}
\caption{Judge=GPT-5.4 bias ($\mathrm{llm\_accuracy} - \mathrm{programmatic\_accuracy}$) by generator, averaged over all difficulty tiers. $^{\ast}$GPT-5.4 judging its own generations.}
\label{tab:self-preference}
\end{table}

GPT-5.4's aggregate self-bias ($+0.012$) falls inside the range spanned by its bias on the four other generators ($-0.024$ to $+0.034$) and is closest to its bias on SmolLM3-3B; it is not an aggregate outlier. The control judge, GPT-OSS-120B, shows a comparable generator-independent bias on GPT-5.4's outputs ($+0.072$) relative to its own cross-generator range ($+0.050$ to $+0.097$), confirming GPT-5.4's generations are not receiving unusual treatment from an unrelated judge either. The one exception is \emph{sequence accuracy}: GPT-5.4-as-judge over-credits its own sequence-accuracy by $+0.172$, versus at most $+0.008$ for any other generator on that same metric under the same judge -- a metric-localised signal we report rather than average away. We do not find evidence of aggregate self-preference, but we cannot rule out a sequence-accuracy-specific effect with this design; a controlled ablation swapping GPT-5.4 out of one role at a time (Limitations) would be needed to isolate the mechanism.

\section{LLM Judge Configurations}
\label{app:judges}

Table~\ref{tab:judges} lists the six LLM judge configurations evaluated in this paper. The set includes large open models (20B-120B), a reasoning-enabled open model (QwQ-32B), and frontier closed models (GPT-5.4, Claude Sonnet 4.5, Gemini-2.5-Pro). QwQ-32B is invoked with \texttt{enable\_thinking=true} (chain-of-thought enabled); all other judges use greedy or near-greedy decoding.

\textbf{Decoding temperature.} vLLM-hosted open judges (QwQ-32B, GPT-OSS-20B, GPT-OSS-120B) use temperature $=0.15$. Frontier judges (GPT-5.4, Claude Sonnet 4.5, Gemini-2.5-Pro) were accessed via provider APIs that did not support temperature $=0$ at time of evaluation; they use the provider's recommended default ($\leq 1.0$). This asymmetry is an implementation constraint, not a design choice. To assess its impact: our temperature sensitivity study (\S\ref{sec:abl:temperature}) shows alignment varies by at most $0.6$\,pp across $T \in \{0.3, 0.7, 1.0\}$, indicating that the inter-condition temperature gap is unlikely to materially confound the cross-judge comparisons. A broader evaluation with harmonised temperatures across all judges is planned for a future revision.

\section{Per-Metric Breakdown}
\label{app:permetric-table}

Table~\ref{tab:e4-permetric} reports per-metric alignment under both
conditions, averaged across all 12 $(g, d)$ configurations. Under with GT,
sequence accuracy shows the widest inter-judge spread: Gemini-2.5-Pro
scores $64.2\%$ while GPT-OSS-20B reaches $86.4\%$, a gap of over 22\,pp.
Tool selection reveals a scale effect: GPT-OSS-20B drops to $70.7\%$,
nearly 18\,pp below GPT-OSS-120B ($88.7\%$). Parameter structure and query
coverage are uniformly high ($86$-$94\%$). Under without-GT, all variation
collapses: the widest spread on any metric is $3.0$\,pp, confirming that
the ``default to $1.0$'' rubric erases capability differences.

\begin{table}[!htbp]
\centering
\small
\setlength{\tabcolsep}{4pt}
\renewcommand{\arraystretch}{1.05}
\resizebox{\columnwidth}{!}{%
\begin{tabular}{@{}lcccc|cccc@{}}
\toprule
 & \multicolumn{4}{c}{\textbf{With GT}} & \multicolumn{4}{c}{\textbf{Without GT}} \\
\cmidrule(lr){2-5}\cmidrule(lr){6-9}
\textbf{Judge}
  & \textbf{tool} & \textbf{param} & \textbf{seq} & \textbf{cov}
  & \textbf{tool} & \textbf{param} & \textbf{seq} & \textbf{cov} \\
\midrule
GPT-5.4         & 83.8          & \textbf{90.0} & 72.3          & \textbf{94.0} & \textbf{84.7} & 89.8 & \textbf{80.7} & 91.0          \\
Claude Sonnet 4.5          & 86.4          & 86.5          & 78.0          & 92.7          & 83.2          & 88.2 & 80.1          & 91.3          \\
Gemini-2.5-Pro  & 82.6          & 89.4          & 64.2          & 93.3          & 82.9          & 89.5 & 80.4          & \textbf{92.3} \\
QwQ-32B         & 88.5          & 89.8          & 84.8          & 93.9          & 82.0          & \textbf{89.9} & 79.9 & 91.5          \\
GPT-OSS-20B     & 70.7          & 89.0          & \textbf{86.4} & 92.5          & 81.7          & 89.2 & 79.5          & 89.5          \\
GPT-OSS-120B    & \textbf{88.7} & 89.7          & 82.9          & 93.5          & 82.1          & \textbf{89.9} & 79.4 & 91.6          \\
\midrule
Prometheus-2    & 71.6          & 43.5          & 70.0          & 56.4          & 67.5          & 49.7 & 73.3  & 60.9           \\
\bottomrule
\end{tabular}%
}
\caption{Per-metric match score $\mu^{m}_{j,r,c} \times 100$, averaged
across all $(g,d)$ configurations. Bold marks the best value per column among
the six main judges. Prometheus-2 (all five generators, both GT conditions) is
shown for reference below the rule and excluded from bold column-wise maxima;
under GT it trails the six-judge range by 20-45\,pp on parameter structure and
query coverage specifically, while remaining roughly competitive on tool
selection and sequence. Without GT, the same pattern holds (parameter
structure and query coverage remain its weakest axes) but every metric moves
in the same direction as its own overall alignment (Table~\ref{tab:full-results}):
tool selection and sequence trade off oppositely, with sequence and query
coverage rising and tool selection falling relative to GT, consistent with a
judge that, lacking a reference to anchor against, defaults to crediting
plausible-looking coverage and ordering while penalising tool selection more
inconsistently.}
\label{tab:e4-permetric}

\end{table}
\begin{table}[h]
\centering
\footnotesize
\setlength{\tabcolsep}{3pt}
\renewcommand{\arraystretch}{0.92}
\begin{tabular*}{\columnwidth}{@{\extracolsep{\fill}}lcccccc@{}}

\toprule

&
\multicolumn{2}{c}{\textbf{Easy}} &
\multicolumn{2}{c}{\textbf{Med.}} &
\multicolumn{2}{c}{\textbf{Hard}} \\

\cmidrule(lr){2-3}
\cmidrule(lr){4-5}
\cmidrule(lr){6-7}

\textbf{Temp.}
& \textbf{GT} & \textbf{w/o}
& \textbf{GT} & \textbf{w/o}
& \textbf{GT} & \textbf{w/o} \\

\midrule

$0.3$
& \textbf{96.4} & \textbf{95.1}
& \textbf{93.5} & \textbf{91.8}
& \textbf{87.5} & \textbf{84.0} \\

$0.7$
& \textbf{96.4} & 95.2
& 93.4 & 91.7
& 87.2 & 84.0 \\

$1.0$
& 96.1 & 95.2
& 92.9 & 91.2
& 87.0 & 83.9 \\

\midrule

\textbf{Spread}
& 0.3 & 0.1
& 0.6 & 0.6
& 0.5 & 0.1 \\

\bottomrule

\end{tabular*}
\caption{Judge temperature sensitivity: per-difficulty alignment (\%) for Qwen3-32B on Llama-3.3-70B.}
\label{tab:app-abl-temp}

\end{table}
\begin{table*}[!htbp]
\centering
\small
\setlength{\tabcolsep}{4pt}
\renewcommand{\arraystretch}{1.05}
\resizebox{0.8\textwidth}{!}{%
\begin{tabular}{@{}llcccccc@{}}
\toprule
 & & \multicolumn{2}{c}{\textbf{Easy}} & \multicolumn{2}{c}{\textbf{Medium}} & \multicolumn{2}{c}{\textbf{Hard}} \\
\cmidrule(lr){3-4}\cmidrule(lr){5-6}\cmidrule(lr){7-8}
\textbf{Generator} & \textbf{Judge} & \textbf{GT} & \textbf{No GT} & \textbf{GT} & \textbf{No GT} & \textbf{GT} & \textbf{No GT} \\
\midrule
\multirow{3}{*}{Llama-3.3-70B}
  & Thinking on  & 96.6 & 95.6 & 93.6 & 92.1 & 87.5 & 84.4 \\
  & Thinking off & 96.5 & 95.6 & 93.6 & 92.0 & 87.3 & 84.4 \\
  & $\Delta$     & $+0.1$ & $0.0$ & $0.0$ & $+0.1$ & $+0.2$ & $0.0$ \\
\midrule
\multirow{3}{*}{Llama-3.1-8B}
  & Thinking on  & 94.2 & 91.2 & 90.8 & 86.8 & 84.1 & 76.6 \\
  & Thinking off & 94.3 & 91.4 & 90.8 & 86.8 & 83.9 & 76.7 \\
  & $\Delta$     & $-0.1$ & $-0.2$ & $0.0$ & $0.0$ & $+0.2$ & $-0.1$ \\
\midrule
\multirow{3}{*}{Qwen3-32B}
  & Thinking on  & 95.0 & 93.9 & 90.9 & 89.1 & 84.7 & 81.1 \\
  & Thinking off & 94.9 & 94.0 & 90.8 & 89.0 & 84.7 & 81.1 \\
  & $\Delta$     & $+0.1$ & $-0.1$ & $+0.1$ & $+0.1$ & $0.0$ & $0.0$ \\
\midrule
\multirow{3}{*}{SmolLM3-3B}
  & Thinking on  & 90.0 & 86.7 & 84.8 & 80.3 & 77.2 & 69.7 \\
  & Thinking off & 89.8 & 86.7 & 84.6 & 80.4 & 76.9 & 69.7 \\
  & $\Delta$     & $+0.2$ & $0.0$ & $+0.2$ & $-0.1$ & $+0.3$ & $0.0$ \\
\bottomrule
\end{tabular}%
}
\caption{Chain-of-thought reasoning study: full per-generator alignment (\%) for QwQ-32B with thinking on vs.\ off. $\Delta$ rows show thinking-on minus thinking-off in percentage points.}
\label{tab:app-abl-e6}

\end{table*}
\section{Supplementary Numerical Tables}
\label{app:ablations}

\subsection*{Judge Temperature: Per-Difficulty Results}

Table~\ref{tab:app-abl-temp} provides the full per-difficulty numerical results for the temperature sensitivity study (Section~\ref{sec:abl:temperature}). Alignment variance across temperatures is $\leq 0.6$\,pp on every (difficulty, condition) slice, confirming that Qwen3-32B judge behaviour is insensitive to sampling stochasticity.

\begin{table}[h!]
\centering
\small
\setlength{\tabcolsep}{4pt}
\renewcommand{\arraystretch}{1.05}
\begin{tabular}{@{}lccc@{}}
\toprule
\textbf{Temperature} & \textbf{Easy} & \textbf{Medium} & \textbf{Hard} \\
\midrule
$0.3$ & 93.19 & 89.73 & 83.52 \\
$0.7$ & 93.05 & 89.93 & 83.57 \\
$1.0$ & 93.07 & 89.69 & 83.59 \\
\midrule
\textbf{Spread} & 0.14 & 0.25 & 0.06 \\
\bottomrule
\end{tabular}
\caption{Judge temperature sensitivity, second pairing: with-GT alignment (\%) for GPT-OSS-120B on Llama-3.1-8B-Instruct. Maximum spread ($0.25$\,pp) is even tighter than the original (Qwen3-32B, Llama-3.3-70B) pairing in Table~\ref{tab:app-abl-temp} ($\leq 0.6$\,pp), confirming temperature insensitivity generalises beyond the original test-bed cell.}
\label{tab:app-abl-temp-r2}
\end{table}

\subsection*{Chain-of-Thought Reasoning: Full Per-Generator Grid}

Table~\ref{tab:app-abl-e6} reports the full $4 \times 3 \times 2$ grid for the QwQ-32B reasoning study (Section~\ref{sec:abl:reasoning}). No cell shows a difference exceeding $0.3$\,pp. The pattern is uniform across generator quality tiers: even on SmolLM3-3B, where generator errors are most frequent and reasoning might be expected to help the judge distinguish correct from incorrect calls, the thinking trace adds nothing.

\section{Per-Topology Breakdown}
\label{app:pertopology-table}

Table~\ref{tab:e5-pertopology} reports per-topology alignment under
with GT. QwQ-32B leads on five of six topologies; GPT-OSS-120B leads on
fan\_out ($93.9\%$) and is a close second elsewhere. Gemini-2.5-Pro is
consistently weakest, trailing QwQ-32B by $5$-$8$\,pp. The gap between
the easiest (fan\_out, $\sim$93\%) and hardest (fan\_in, $\sim$83\%)
topologies is roughly 10\,pp, comparable to the easy-to-hard difficulty
degradation in Table~\ref{tab:full-results}. Fan-out's advantage is
intuitive: each parallel branch can be verified independently, whereas
fan-in and loop-like topologies require tracking cross-branch dependencies.

\begin{table}[!htbp]
\centering
\small
\setlength{\tabcolsep}{4pt}
\renewcommand{\arraystretch}{1.05}
\resizebox{\columnwidth}{!}{%
\begin{tabular}{@{}lcccccc@{}}
\toprule
\textbf{Judge} & \textbf{linear} & \textbf{fan\_out} & \textbf{fan\_in} & \textbf{diamond} & \textbf{opt.\ enrich} & \textbf{loop\_like} \\
\midrule
GPT-5.4         & 84.3          & 92.6          & 83.5          & 83.1          & 85.5          & 83.0          \\
Claude Sonnet 4.5          & 84.4          & 92.4          & 83.7          & 86.9          & 86.5          & 85.2          \\
Gemini-2.5-Pro  & 80.8          & 91.7          & 80.0          & 82.0          & 82.9          & 81.1          \\
QwQ-32B         & \textbf{88.1} & 93.3          & \textbf{87.8} & \textbf{90.2} & \textbf{90.1} & \textbf{87.7} \\
GPT-OSS-20B     & 83.5          & 92.7          & 82.5          & 84.0          & 85.2          & 83.0          \\
GPT-OSS-120B    & 87.5          & \textbf{93.9} & 87.0          & 89.4          & 89.4          & 87.1          \\
\bottomrule
\end{tabular}%
}
\caption{Judge alignment (with GT, \%) by DAG topology. Bold marks the
best judge per topology.}
\label{tab:e5-pertopology}

\end{table}

\begin{figure}[!htbp]
\centering
\includegraphics[width=0.9\columnwidth]{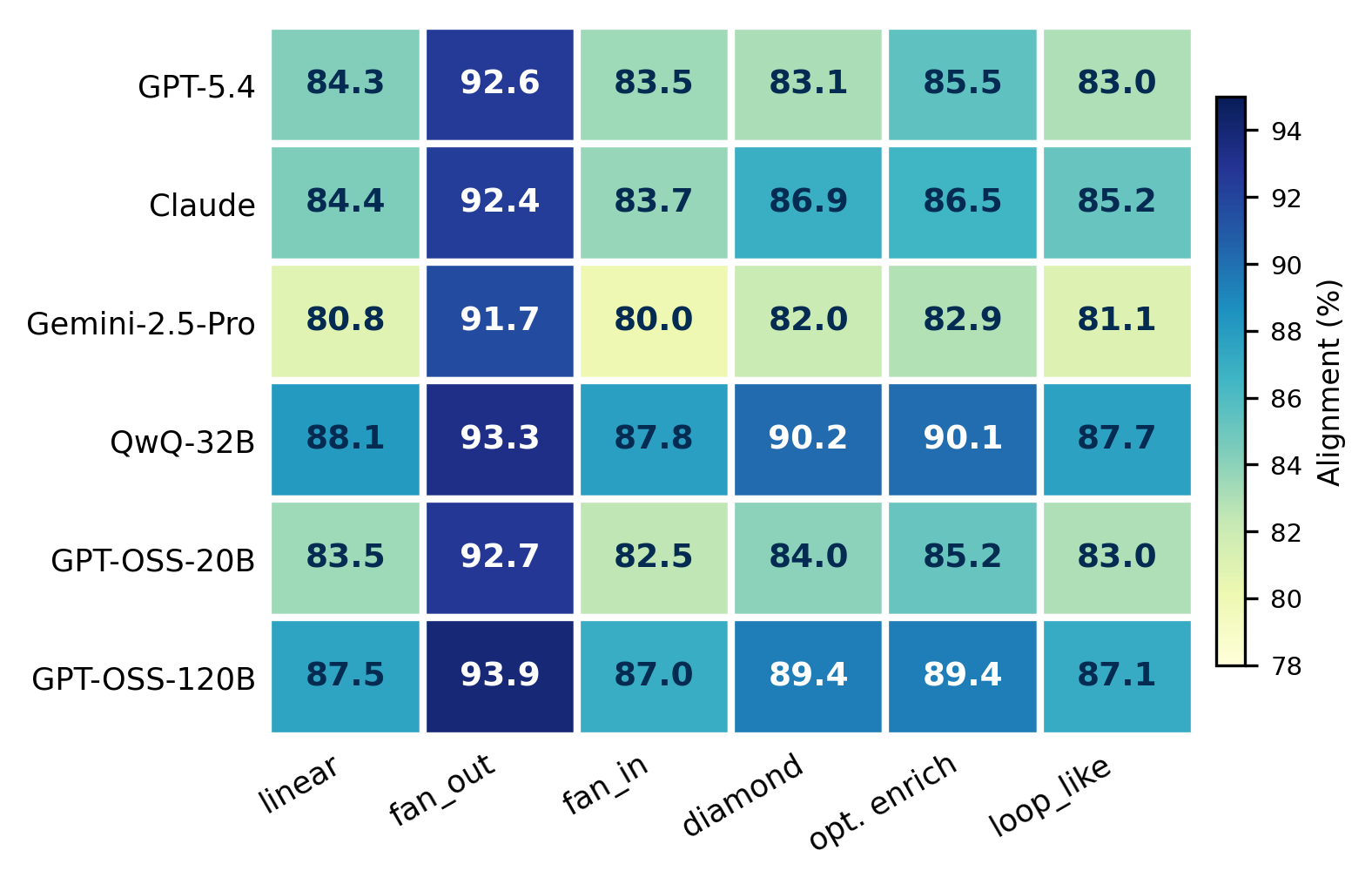}
\caption{Judge alignment (\%) by DAG topology (with GT), visualised as a heatmap. Fan-out consistently achieves the highest alignment across all judges; fan-in and loop-like are the hardest. The ordering is judge-independent, indicating that DAG structural complexity is an intrinsic difficulty signal.}
\label{fig:topo-heatmap}
\end{figure}

\section{Inter-Judge Confusion Matrices}
\label{app:confusion}

Table~\ref{tab:confusion-full} reports per-verdict counts for the highest-agreement pair (QwQ-32B$\times$GPT-OSS-120B, $\kappa=0.606$) and the lowest-agreement pair (Claude Sonnet 4.5$\times$GPT-OSS-20B, $\kappa=0.225$) under with GT. Disagreement concentrates almost entirely at the partial-credit boundary: $94.7\%$ and $96.8\%$ of off-diagonal entries involve at least one $0.5$ verdict. The dominant off-diagonal cell for the low-agreement pair is Claude Sonnet 4.5$=1$ / GPT-OSS-20B$=0.5$ ($18.2\%$ of verdicts), indicating GPT-OSS-20B is systematically more conservative. Under without GT, both pairs converge as both judges default to $1.0$: Claude Sonnet 4.5$\times$OSS-20B agreement rises from $70.3\%$ to $90.1\%$.

\begin{figure}[!htbp]
\centering
\small
\setlength{\tabcolsep}{4pt}
\renewcommand{\arraystretch}{1.1}

\begin{minipage}[t]{0.45\linewidth}
\centering
\textbf{QwQ-32B $\times$ GPT-OSS-120B}\\
{\scriptsize ($\kappa=0.61$, Agreement 87.9\%)}

\vspace{4pt}

\begin{tabular}{c|ccc}
\toprule
 & \textbf{0} & \textbf{0.5} & \textbf{1} \\
\midrule
\textbf{0}   & \cellcolor{gray!15}\textbf{1.3} & 0.8 & 0.3 \\
\textbf{0.5} & 1.3 & \cellcolor{gray!15}\textbf{10.3} & 4.9 \\
\textbf{1}   & 0.4 & 4.4 & \cellcolor{gray!15}\textbf{76.3} \\
\bottomrule
\end{tabular}
\end{minipage}
\hfill
\begin{minipage}[t]{0.45\linewidth}
\centering
\textbf{Claude Sonnet 4.5 $\times$ GPT-OSS-20B}\\
{\scriptsize ($\kappa=0.23$, Agreement 70.3\%)}

\vspace{4pt}

\begin{tabular}{c|ccc}
\toprule
 & \textbf{0} & \textbf{0.5} & \textbf{1} \\
\midrule
\textbf{0}   & \cellcolor{gray!15}\textbf{0.7} & 0.4 & 0.2 \\
\textbf{0.5} & 1.9 & \cellcolor{gray!15}\textbf{9.7} & 8.3 \\
\textbf{1}   & 0.7 & 18.2 & \cellcolor{gray!15}\textbf{60.0} \\
\bottomrule
\end{tabular}
\end{minipage}

\captionof{figure}{Verdict-level confusion matrices (\% of $N$). Rows = judge 1, columns = judge 2. Diagonal entries are shaded.}
\label{tab:confusion-full}
\end{figure}

\subsection*{Binary vs.\ \{0, 0.5, 1\} Verdict Scale}
\label{app:abl-binary}

Table~\ref{tab:abl-binary} compares judge rankings under the original three-level $\{0, 0.5, 1\}$ verdict scale vs.\ a binary collapse in which each $0.5$ verdict is remapped to $0$ if the programmatic reference is below $0.5$, and $1$ otherwise. Columns report mean GT alignment across all 15 (generator, difficulty) cells per judge, and the per-judge rate of issuing $0.5$ verdicts.

\begin{table}[t]
\centering
\footnotesize
\setlength{\tabcolsep}{3.5pt}
\renewcommand{\arraystretch}{0.95}

\begin{tabular*}{\columnwidth}{@{\extracolsep{\fill}}lcccc@{}}

\toprule

\textbf{Judge}
& \textbf{Std.}
& \textbf{Binary}
& \textbf{$\Delta$}
& \textbf{0.5 rate} \\

\midrule

GPT-OSS-20B
& 84.0 & \textbf{97.0} & $+13.0$ & 26.5\% \\

QwQ-32B
& \textbf{87.6} & 94.7 & $+7.1$ & 15.2\% \\

GPT-OSS-120B
& 87.5 & 95.4 & $+7.9$ & 14.0\% \\

Claude Sonnet 4.5$^{\ddagger}$
& 85.4 & 94.7 & $+9.2$ & 18.6\% \\

Gemini-2.5-Pro
& 81.7 & 90.0 & $+8.3$ & 12.2\% \\

GPT-5.4
& 84.8 & \textbf{96.6} & $+11.9$ & 28.2\% \\

\bottomrule

\end{tabular*}
\caption{Binary vs.\ $\{0,0.5,1\}$ verdict scale: mean GT alignment (\%) and judge rankings averaged over 15 (generator, difficulty) cells. $\Delta$ = Binary $-$ Standard.}
\label{tab:abl-binary}

\end{table}

The binary collapse does not inflate all scores uniformly: judges with high $0.5$ rates (GPT-5.4: $28.2\%$, GPT-OSS-20B: $26.5\%$) gain disproportionately ($+11.9$\,pp and $+13.0$\,pp respectively). Under binary remapping, the $0.5$ verdict is always counted as correct because it is mapped to the programmatic direction by construction. Judges that hedge on uncertain records therefore receive artificially inflated binary alignment, reshuffling the ranking from QwQ-32B $>$ GPT-OSS-120B to GPT-OSS-20B $>$ GPT-5.4. The Spearman rank correlation between standard and binary ranking is $\rho = 0.03$ ($p = 0.96$), indicating the rankings are essentially uncorrelated. This confirms that the $0.5$ verdict is not random noise: it encodes directional uncertainty that is systematically correlated with judge capacity and lost under binary collapse. We recommend retaining the three-level scale.

\section{Judge Score Stochasticity and Prompt Sensitivity}
\label{app:stochasticity}

\begin{table*}[h!]
\centering
\small
\renewcommand{\arraystretch}{0.95}
\setlength{\tabcolsep}{4pt}
\resizebox{0.85\textwidth}{!}{%
\begin{tabular}{l l p{6cm}}
\toprule
\textbf{Name} & \textbf{Year} & \textbf{Description} \\
\midrule
\multicolumn{3}{l}{\textit{Agentic Data \& Tool-Calling Benchmarks}} \\
\midrule
Self-Instruct \citep{wang2023selfinstruct} & 2022 & LLM-only instruction synthesis; seed for synthetic data pipelines. \\
Mind2Web \citep{deng2023mind2web} & 2023 & Human-annotated user trajectories over real websites. \\
AgentBench \citep{liu2024agentbench} & 2023 & Multi-domain agentic benchmark across reasoning and interaction settings. \\
WebArena \citep{zhou2024webarena} & 2024 & Realistic web environment with functional sites and long-horizon tasks. \\
WorkArena \citep{drouin2024workarena} & 2024 & Evaluation on production websites for task-solving agents. \\
AppWorld \citep{trivedi2024appworld} & 2024 & Multi-app environment for interactive code-based agent workflows. \\
TaskBench \citep{shen2024taskbench} & 2024 & Tool-graph-based benchmark for task decomposition and tool selection. \\
AgentTrek \citep{xu2025agenttrek} & 2025 & Converts web tutorials into executable agent trajectories. \\
BFCL \citep{patil2025bfcl} & 2025 & Leaderboard for single-turn function-calling accuracy; AST-based scoring. \\
FuncBenchGen \citep{maekawa2025funcbenchgen} & 2025 & DAG-based synthetic function-calling with controllable complexity. \\
$\tau$-bench \citep{yao2024taubench} & 2024 & Tool-agent-user interaction with simulated users and \texttt{pass\textasciicircum k} reliability metric. \\
\midrule
\multicolumn{3}{l}{\textit{LLM-as-Judge}} \\
\midrule
MT-Bench \citep{zheng2023judging} & 2023 & Established the LLM-as-judge paradigm; identified positional, verbosity, and self-enhancement biases. \\
JudgeBench \citep{tan2025judgebench} & 2024 & Judge evaluation on hard pairs with verifiable ground truth. \\
Arena-Hard-Auto \citep{li2024arenahard} & 2024 & Automated pairwise judge evaluation; high human agreement. \\
PandaLM \citep{wang2024pandalm} & 2024 & Dedicated judge model trained for pairwise comparison. \\
Auto-J \citep{li2024autoj} & 2024 & Generalist judge model trained on diverse evaluation criteria. \\
JudgeLM \citep{zhu2023judgelm} & 2023 & Fine-tuned 7B--33B judges; characterises position, knowledge, and format biases. \\
Prometheus 2 \citep{kim2024prometheus2} & 2024 & Open-weight evaluator with rubric-conditioned direct and pairwise assessment. \\
\midrule
\multicolumn{3}{l}{\textit{LLM Judges for Tool-Calling Evaluation}} \\
\midrule
ToolEval \citep{qin2024toolllm} & 2023 & ChatGPT as judge for pass rate on API trajectories. \\
StableToolBench \citep{guo2024stabletoolbench} & 2024 & GPT-4-turbo as evaluator in a virtualized API environment. \\
ToolSandbox \citep{lu2024toolsandbox} & 2024 & Questioned LLM judge reliability; replaced with milestone-based programmatic scoring. \\
MCP-AgentBench \citep{guo2025mcpagentbench} & 2025 & Hybrid rule-based and LLM judge for task-completion scoring. \\
GeoBenchX \citep{krechetova2025geobenchx} & 2025 & Three-judge panel for geospatial tool-use evaluation. \\
Agent-as-a-Judge \citep{zhuge2024agentjudge} & 2024 & Agent evaluates another agent on DAG-structured development tasks. \\
Auto-Eval Judge \citep{bhonsle2025autoeval} & 2025 & Modular framework decomposing evaluation into checklist questions. \\
\bottomrule
\end{tabular}%
}
\caption{Representative works across agentic benchmarks, LLM-as-judge methods, and LLM judges for tool-calling evaluation.}
\label{tab:overview}

\end{table*}
\subsection*{Repeated-run variability}

A natural concern for any LLM-based evaluation pipeline is whether the judge's score for a given record is stable across independent runs with the same prompt. All frontier judges (GPT-5.4, Claude Sonnet~4.5, Gemini-2.5-Pro) were called at near-default temperatures ($\leq 1.0$, as noted in Appendix~\ref{app:judges}), meaning a small amount of stochasticity is inherent to each call. Our temperature sensitivity study (Section~\ref{sec:abl:temperature} and Appendix~\ref{app:ablations}) indirectly characterises this: Qwen3-32B alignment varies by at most $0.6$\,pp across $T \in \{0.3, 0.7, 1.0\}$ on 3,771 records, providing an upper bound on within-judge run-to-run variance. We therefore expect configuration-level alignment estimates (averaged over $N_{g,d} \geq 3{,}764$ records) to be highly stable; a $0.6$\,pp spread at record level contracts to ${\ll}0.1$\,pp at the configuration mean by the central limit theorem.

\subsection*{Prompt variation sensitivity}

Our prompt ablation (Section~\ref{sec:abl:format}, Figure~\ref{fig:abl-format}) shows that switching from the structured per-metric JSON rubric to a free-form one-sentence instruction drops GT alignment by $4.8$--$6.5$\,pp for Qwen3-32B on the Llama-3.3-70B generator. This $>5$\,pp gap on the original pairing dwarfs the within-prompt stochasticity bound and confirms that prompt structure is a substantial source of judge-score variability on that pairing, not sampling noise. Practitioners adapting these prompts should expect similar sensitivity: minor wording changes (e.g., removing the ``default to 1.0'' instruction) can shift alignment by $1$--$5$\,pp, as demonstrated by the C2 ablation (Table~\ref{tab:c2-ceiling}).

We tested whether the format effect generalises on a second pairing, QwQ-32B on SmolLM3-3B (Table~\ref{tab:app-abl-format-r2}). The direction replicates on easy ($+3.9$\,pp) and medium ($+2.4$\,pp) but is smaller than on the original pairing, and \emph{reverses} on hard ($-0.8$\,pp: free-form marginally ahead). We therefore revise our characterisation: prompt format is not a uniformly dominant lever independent of judge, generator, or difficulty: it is a real and sometimes large effect, but its magnitude and even its direction on hard queries depend on the specific pairing (see Limitations). Standalone prompt texts for all four variants used in this study are provided in Appendix~\ref{app:prompts}.

\begin{table}[h!]
\centering
\small
\setlength{\tabcolsep}{4pt}
\renewcommand{\arraystretch}{1.05}
\begin{tabular}{@{}lccc@{}}
\toprule
 & \textbf{Easy} & \textbf{Medium} & \textbf{Hard} \\
\midrule
Structured & 89.98 & 84.78 & 77.22 \\
Free-form  & 86.09 & 82.43 & 78.05 \\
\midrule
\textbf{$\Delta$ (Struct.$-$Free)} & $+3.88$ & $+2.35$ & $-0.83$ \\
\bottomrule
\end{tabular}
\caption{Prompt format ablation, second pairing: with-GT alignment (\%) for QwQ-32B on SmolLM3-3B, structured per-metric prompt vs.\ free-form. Compare to the original (Qwen3-32B, Llama-3.3-70B) pairing's $+4.8$-$+6.5$\,pp (Figure~\ref{fig:abl-format}): the effect is smaller here and reverses sign on hard queries.}
\label{tab:app-abl-format-r2}
\end{table}

\section{Related Work Survey}
\label{app:relwork-table}
A rigorous per-judge repeated-run study is planned for a future revision: running each judge twice on a stratified subset and computing per-record verdict-flip rates. Such a study would directly quantify the fraction of borderline verdicts driven by stochasticity rather than systematic judge disagreement, and would allow score variance to be separated from the inter-judge disagreement reported in Table~\ref{tab:interjudge}.

\section{Prompts}
\label{app:prompts}
\label{app:judge-prompts}

This appendix reproduces, verbatim, every prompt used in the evaluation pipeline: (i)~the \emph{generator} prompt (Appendix~\ref{app:prompt-generator}) that instructs each generator $g \in \mathcal{G}$ to emit a tool-call sequence, (ii)~the \emph{with GT judge} prompt (Appendix~\ref{app:prompt-gt}) sent to every LLM judge $j \in \mathcal{J}$ under the GT condition, (iii)~the \emph{without GT judge} prompt (Appendix~\ref{app:prompt-without GT}), identical to (ii) save for the omission of the \textsc{Expected Tool Calls} block, and (iv)~the \emph{Prometheus-2} judge prompt (Appendix~\ref{app:prompt-prometheus2}), following the model's official absolute-grading template. Decoding parameters are held constant across all prompts and all $(g, j, d, c)$ configurations.

\subsection{Generator prompt.}
\label{app:prompt-generator}

\begin{promptbox}[\textsc{System}]
\ttfamily\small
You are a tool-call generation model participating in a BENCHMARK EVALUATION.\\[2pt]
This is a SIMULATION. The tools listed are HYPOTHETICAL. You are NOT being asked to actually execute anything. You are being evaluated on your ability to select and format the correct tool calls given a query and a tool schema.\\[4pt]
\textbf{Your ONLY job:}\\
Given the user message and available tools, output the correct JSON array of tool calls. Nothing else.\\[4pt]
\textbf{Rules:}\\
- Output ONLY a raw JSON array. No text, no markdown, no explanation.\\
- If no tools apply, output: \texttt{[]}\\
- Use EXACTLY the format specified below.\\[4pt]
\textit{\{system\_prompt\}}\\[4pt]
\textbf{Available Tools:}\\
\textit{\{available\_tools\}}
\end{promptbox}

\begin{promptbox}[\textsc{User}]
\ttfamily\small
\textit{\{user\_message\}}
\end{promptbox}

\subsection{Judge prompt: with ground truth (GT condition).}
\label{app:prompt-gt}

\begin{promptbox}[\textsc{User}]
\ttfamily\small
You are a STRICT STRUCTURAL EVALUATOR for AI agent tool-calling plans.\\[2pt]
IMPORTANT CONTEXT (NON-NEGOTIABLE):\\
- The evaluated model is SINGLE-SHOT and STATELESS.\\
- The model does NOT observe tool execution or tool outputs.\\
- The model ONLY plans tool calls.\\
- ALL evaluation MUST be STATIC and STRUCTURAL.\\
- \textbf{Assume the programmatic judge is the ground truth.}\\[2pt]
ABSOLUTELY DO NOT:\\
- Evaluate correctness of parameter VALUES\\
- Compare generated values to expected values\\
- Infer correctness from tool output logic\\
- Penalize placeholder, symbolic, or dummy values\\
- Penalize derived-value mismatches (e.g., 0.75 vs 0.005)\\[2pt]
ONLY evaluate:\\
- Tool presence\\
- Tool names\\
- Parameter NAMES\\
- Parameter STRUCTURE (dict shape)\\
- Logical ordering of calls\\
- Coverage of user intent\\[2pt]
If parameter NAMES match and the tool intent is satisfied, the parameter structure MUST be treated as correct.\\[4pt]
-\\
\textbf{Evaluation Metrics (PROGRAMMATIC-ALIGNED)}\\
You must compute FOUR scores. Scores should default to \textbf{1.0 unless a clear structural violation exists}.\\[4pt]
\textbf{1. TOOL SELECTION ACCURACY.} Evaluate only: required tools present, no clearly irrelevant tools. Ignore tool grouping and execution feasibility. Scoring: 1.0 all required/no irrelevant; 0.5 missing or clearly irrelevant; 0.0 plan fundamentally incorrect.\\[3pt]
\textbf{2. PARAMETER STRUCTURE ACCURACY (MOST IMPORTANT).} Evaluate strictly: presence of required parameter names; arguments are a dictionary; parameter names logically belong to the tool. Do NOT evaluate parameter values, numeric ranges, threshold correctness, dependency resolution, or output-derived values. Rules: required parameter names present $\Rightarrow$ structurally correct; optional parameters never reduce score; placeholder or symbolic values are always valid. Scoring: 1.0 correct names on all tools; 0.5 minor missing/extra; 0.0 required names missing for most tools.\\[3pt]
\textbf{3. SEQUENCE \& DEPENDENCY ACCURACY.} Does the order reflect logical planning? Are prerequisites placed earlier? Assume symbolic dependency handling is valid. Scoring: 1.0 logical; 0.5 minor ordering issue, intent preserved; 0.0 illogical or contradictory.\\[3pt]
\textbf{4. QUERY COVERAGE ACCURACY.} Does the plan cover all parts of the user request? Are all sub-goals planned for? Scoring: 1.0 full coverage; 0.5 partial; 0.0 major intent missed.\\[4pt]
-\\
\textbf{OUTPUT FORMAT (STRICT).} Return exactly one JSON object with the four metric fields, each containing \texttt{\{"accuracy": <float>, "justification": "<1 sentence>"\}}, plus an \texttt{"overall\_assessment"} single-sentence summary.\\[4pt]
Be conservative. Prefer 1.0 unless a \textbf{clear structural violation} exists. Judge structure, not semantics.\\[6pt]
QUERY:\\
\textit{\{user\_message\}}\\[4pt]
AVAILABLE TOOLS:\\
\textit{\{available\_tools\}}\\[4pt]
GENERATED TOOL CALLS:\\
\textit{\{generated\_tool\_calls\}}\\[4pt]
EXPECTED TOOL CALLS:\\
\textit{\{expected\_responses\}}
\end{promptbox}

\subsection{Judge prompt: without ground truth (without-GT condition).}
\label{app:prompt-without GT}

\begin{promptbox}[\textsc{User}]
\ttfamily\small
You are a STRICT STRUCTURAL EVALUATOR for AI agent tool-calling plans.\\[2pt]
IMPORTANT CONTEXT (NON-NEGOTIABLE):\\
- The evaluated model is SINGLE-SHOT and STATELESS.\\
- The model does NOT observe tool execution or tool outputs.\\
- The model ONLY plans tool calls.\\
- ALL evaluation MUST be STATIC and STRUCTURAL.\\
- There is NO expected tool plan available.\\[2pt]
ABSOLUTELY DO NOT:\\
- Evaluate correctness of parameter VALUES\\
- Infer correctness from tool output logic\\
- Penalize placeholder, symbolic, or dummy values\\
- Penalize derived-value mismatches\\
- Penalize missing runtime-dependent values\\[2pt]
ONLY evaluate using:\\
- The user QUERY\\
- AVAILABLE TOOLS schema\\
- GENERATED TOOL CALLS\\[2pt]
Treat the AVAILABLE TOOLS as the ONLY source of truth.\\[4pt]
-\\
\textbf{Evaluation Metrics (STRUCTURAL ONLY)}\\
Compute FOUR scores. Scores MUST default to \textbf{1.0 unless a clear structural violation exists}.\\[4pt]
\textbf{1. TOOL SELECTION ACCURACY.} Are selected tools relevant to the query? Tools that clearly do not belong? Rules: if a tool plausibly helps answer the query, treat as valid; only penalize clearly irrelevant tools. Scoring: 1.0 all relevant; 0.5 one questionable/missing; 0.0 selection unrelated.\\[3pt]
\textbf{2. PARAMETER STRUCTURE ACCURACY (MOST IMPORTANT).} Evaluate strictly: arguments exist and are dictionaries; parameter names exist in the tool schema; parameter names logically belong to the tool. Do NOT evaluate parameter values, types, ranges, thresholds, or symbolic references. Rules: matching names $\Rightarrow$ correct; optional parameters never reduce score; extra parameters only penalize if clearly invalid. Scoring: 1.0 valid across all tools; 0.5 minor issues; 0.0 most tools incorrect.\\[3pt]
\textbf{3. SEQUENCE \& DEPENDENCY ACCURACY.} Is ordering logically consistent with the query? Are analysis steps before decision/recommendation steps? Assume symbolic dependencies are valid. Scoring: 1.0 logical; 0.5 minor issue; 0.0 illogical.\\[3pt]
\textbf{4. QUERY COVERAGE ACCURACY.} Does the plan attempt to address all parts of the query? Are key sub-goals represented? Scoring: 1.0 full; 0.5 partial; 0.0 major intent missed.\\[4pt]
-\\
\textbf{OUTPUT FORMAT (STRICT).} Return exactly one JSON object with the four metric fields, each containing \texttt{\{"accuracy": <float>, "justification": "<1 sentence>"\}}, plus an \texttt{"overall\_assessment"} single-sentence summary.\\[4pt]
Be conservative. Prefer 1.0 unless a \textbf{clear structural violation} exists. Judge structure, not execution or correctness.\\[6pt]
QUERY:\\
\textit{\{user\_message\}}\\[4pt]
AVAILABLE TOOLS:\\
\textit{\{available\_tools\}}\\[4pt]
GENERATED TOOL CALLS:\\
\textit{\{generated\_tool\_calls\}}
\end{promptbox}

\subsection{Judge prompt: Prometheus-2 (judge-specialised baseline).}
\label{app:prompt-prometheus2}

Prometheus-2 \citep{kim2024prometheus2} is fine-tuned for single-metric absolute grading and cannot score four metrics in one call; we follow the model's official absolute-grading template and system message, issuing four independent calls per record (one per metric, differing only in the score rubric block), then merge the four verdicts into the same judge-response schema every other judge produces.

\begin{promptbox}[\textsc{System}]
\ttfamily\small
You are a fair judge assistant tasked with providing clear, objective feedback based on specific criteria, ensuring each assessment reflects the absolute standards set for performance.
\end{promptbox}

\begin{promptbox}[\textsc{User}]
\ttfamily\small
\#\#\#Task Description:\\
An instruction (an agentic tool-calling query with its available tool schemas), a response to evaluate (a proposed tool-call plan), a reference answer that gets a score of 5, and a score rubric representing an evaluation criteria are given.\\
1. Write a detailed feedback that assesses the quality of the response strictly based on the given score rubric, not evaluating in general.\\
2. After writing a feedback, write a score that is an integer between 1 and 5. You should refer to the score rubric.\\
3. The output format should look as follows: "Feedback: (write a feedback for criteria) [RESULT] (an integer number between 1 and 5)"\\
4. Please do not generate any other opening, closing, and explanations.\\[4pt]
\#\#\#The instruction to evaluate:\\
Given the user query and available tool schemas below, select and structure the correct tool call(s) to satisfy the query.\\[2pt]
QUERY:\\
\textit{\{user\_message\}}\\[4pt]
AVAILABLE TOOLS:\\
\textit{\{available\_tools\}}\\[4pt]
\#\#\#Response to evaluate:\\
\textit{\{generated\_tool\_calls\}}\\[4pt]
\#\#\#Reference Answer (Score 5):\\
\textit{\{expected\_responses\}}\\[4pt]
\#\#\#Score Rubrics:\\
\textit{\{one of four metric-specific rubrics: tool selection, parameter structure, sequence \& dependency, or query coverage -- each a bracketed criterion followed by five score descriptions, structurally identical to the with GT judge rubric in Appendix~\ref{app:prompt-gt}\}}\\[4pt]
\#\#\#Feedback:
\end{promptbox}

\section{Free-Form Judge Prompt (A5)}
\label{app:prompt-freeform}

The following prompt is used in the A5 prompt-format ablation (\S\ref{sec:abl:format}). It omits explicit per-metric definitions and scoring rubrics, asking the judge to reason freely and return a single holistic verdict per metric without anchoring instructions.

\begin{promptbox}[\textsc{User}]
\ttfamily\small
You are evaluating an AI agent's tool-calling plan.\\[4pt]
Given the user query, the available tools, and the generated tool calls, assess how well the agent performed. Consider whether it selected appropriate tools, structured parameters correctly, ordered calls logically, and covered the user's intent.\\[4pt]
Return a JSON object with four keys: \texttt{tool\_selection\_accuracy}, \texttt{parameter\_structure\_accuracy}, \texttt{sequence\_accuracy}, and \texttt{query\_coverage\_accuracy}. Each key should map to an object with \texttt{"accuracy"} (a float between 0 and 1) and \texttt{"justification"} (one sentence). Also include an \texttt{"overall\_assessment"} field.\\[6pt]
QUERY:\\
\textit{\{user\_message\}}\\[4pt]
AVAILABLE TOOLS:\\
\textit{\{available\_tools\}}\\[4pt]
GENERATED TOOL CALLS:\\
\textit{\{generated\_tool\_calls\}}\\[4pt]
EXPECTED TOOL CALLS:\\
\textit{\{expected\_responses\}}
\end{promptbox}

\end{document}